\documentclass[10pt,journal,letterpaper,compsoc]{IEEEtran}

\ifCLASSOPTIONcompsoc
\else
\fi
\ifCLASSINFOpdf
\else
\fi
\usepackage{algorithm}
\usepackage{algorithmicx}
\usepackage{algpseudocode}
\usepackage{amsmath}
\usepackage{graphics}
\usepackage{epsfig}
\usepackage{subfigure}
\usepackage{graphicx}
\usepackage{amssymb}
\usepackage{threeparttable}
\usepackage[usenames, dvipsnames]{color}
\usepackage[normalem]{ulem}
\usepackage{multirow}
\usepackage{float}
\usepackage{amsfonts}
\usepackage{bm}
\usepackage{array}
\usepackage[table]{xcolor}
\usepackage{colortbl}
\usepackage{ragged2e}
\usepackage{lineno}

\usepackage[breaklinks=true,letterpaper=true,colorlinks,bookmarks=false]{hyperref}

\makeatletter 
\newcommand{\Rmnum}[1]{\expandafter\@slowromancap\romannumeral #1@}
\makeatother

\begin{document}
	\title{Distilled Siamese Networks for Visual Tracking}
	\author{Jianbing Shen,~\IEEEmembership{Senior Member,~IEEE}, Yuanpei Liu, Xingping Dong, Xiankai Lu, \\
     Fahad Shahbaz Khan, and Steven Hoi,~\IEEEmembership{Fellow,~IEEE}
		\IEEEcompsocitemizethanks{
			\IEEEcompsocthanksitem Jianbing Shen, Xingping Dong, and Fahad Shahbaz Khan are
             with the Inception Institute of Artificial Intelligence, Abu Dhabi, UAE.
              (Email: shenjianbingcg@gmail.com)
			\IEEEcompsocthanksitem  Yuanpei Liu is with School of Computer Science,
			Beijing Institute of Technology, Beijing 100081, P. R. China.
			\IEEEcompsocthanksitem  Xiankai Lu is with School of Software,
	        Shandong University, Jinan, P. R. China.
            \IEEEcompsocthanksitem Steven Hoi is with the School of Information Systems,
            Singapore Management University, and Salesforce Research Asia, Singapore.
		}
	}
	
	\markboth{IEEE Transactions on Pattern Analysis and Machine Intelligence} 
	{Shell \MakeLowercase{\textit{et al.}}: Bare Demo of IEEEtran.cls
		for Computer Society Journals}
	%
	
	
	
	\IEEEcompsoctitleabstractindextext{%
\begin{abstract} In recent years, Siamese network based trackers have significantly advanced the state-of-the-art in real-time tracking. Despite their success, Siamese trackers tend to suffer from high memory costs, which restrict their applicability to mobile devices with tight memory budgets. To address this issue, we propose a distilled Siamese tracking framework to learn small, fast and accurate trackers (students), which capture critical knowledge from large Siamese trackers (teachers) by a teacher-students knowledge distillation model. This model is intuitively inspired by the one teacher vs. multiple students learning method typically employed in schools. In particular, our model contains a single teacher-student distillation module and a student-student knowledge sharing mechanism. The former is designed using a tracking-specific distillation strategy to transfer knowledge from a teacher to students. The latter is utilized for mutual learning between students to enable in-depth knowledge understanding. Extensive empirical evaluations on several popular Siamese trackers demonstrate the generality and effectiveness of our framework. Moreover, the results on five tracking benchmarks show that the proposed distilled trackers achieve compression rates of up to 18$\times$ and frame-rates of $265$ FPS, while obtaining {comparable tracking accuracy compared to base models.}
\end{abstract}
		
		}
	
	\maketitle
	\IEEEdisplaynotcompsoctitleabstractindextext

	\IEEEpeerreviewmaketitle
	\begin{figure}
		\centering
		\includegraphics[width = 0.5 \textwidth]{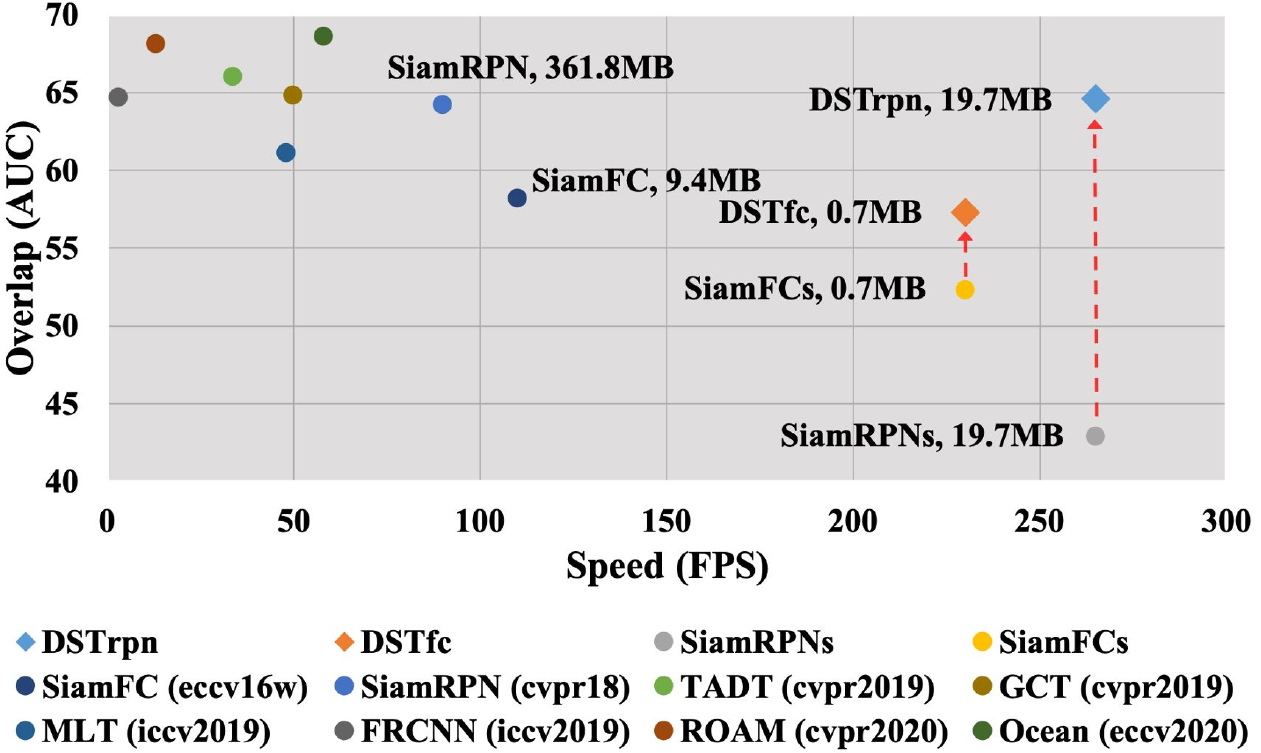}
		\caption{\textbf{Comparison in terms of speed (FPS) and accuracy (AUC) of state-of-the-art (SOTA) trackers on OTB-100~\cite{wuobject2015},} including SiamRPN~\cite{li2018high}, SiamFC~\cite{bertinetto2016fully}, TADT~\cite{li2019target}, GCT~\cite{gao2019graph}, MLT~\cite{choi2019deep}, FRCNN~\cite{huang2019bridging}, ROAM~\cite{yang2020roam}, and Ocean~\cite{zhang2020ocean}.
		{Compared with the small models trained from scratch (SiamRPNs and SiamFCs), our models (DSTrpn and DSTfc) trained with the knowledge distillation method show significant improvements.}
		Further, DSTrpn achieves a 3$\times$~speed, 18$\times$~memory compression rate {and comparable accuracy over its teacher variant} (SiamRPN~\cite{li2018high}).
		Besides, both DSTrpn (SiamRPN~\cite{li2018high} as teacher) and DSTfc (SiamFC~\cite{bertinetto2016fully} as teacher) obtain competitive accuracy while achieving the highest speed.}
		\label{fig:example}
	\end{figure}

	\section{Introduction}		
	Recent years have witnessed a significant increase {in Siamese-based tracking methods} due to their strong balance between accuracy and speed.
	{The pioneering work, SINT~\cite{tao2016siamese}, first introduced Siamese network to the visual tracking community, achieving promising tracking performance.
	Afterwards, SiamFC~\cite{bertinetto2016fully} introduced a simple yet effective tracking framework that makes use of offline training to learn a metric function. This learned metric is then utilized to convert the tracking task to that of template matching.}
    This framework serves as an ideal {\it baseline} for real-time tracking due to its simple architecture and high speed.
    As such, many real-time trackers~\cite{valmadre2017end,dong2018hyperparameter,dong2018triplet,li2018high,zhu2018distractor, Dong2021hyper, gao2019graph, choi2019deep,Lu2020Shrinkage} have been proposed as the extension of SiamFC, with improved accuracy. For instance, the recent tracker SiamRPN~\cite{li2018high, Kristan2018a} (champion of the VOT-2018~\cite{Kristan2018a} challenge), achieved significantly improved accuracy and high speed (nearly 90 FPS), by applying a Region Proposal Network (RPN) to directly regress the position and scale of objects. This method has the potential to likely become the next {\it baseline} for further promoting real-time tracking, due to its high speed and impressive accuracy.
	
	Despite being actively studied with remarkable progress, Siamese-network based visual trackers generally face a conflict between their high memory cost and the strict constraints on memory budget in real-world applications, especially in case of
	SiamRPN~\cite{li2018high,Kristan2018a} whose model size is up to 361.8 MB.
	In particular, the high memory costs of current trackers make them undesirable for important application in mobile devices, such as smartphones.
	{One simple solution is directly reducing the model size of a tracker and then training the small model with the original method, however, it often leads to dramatically drop of performance. For instance, we retrain the small models of two representative Siamese trackers: SiamRPN and SiamFC. As shown in Fig.~\ref{fig:example}, the retrained models: SiamRPNs and SiamFCs suffer from large performance degradation with original training methods.}
	Thus, developing a way to reduce the memory cost of Siamese trackers without a significant loss in tracking accuracy is essential in order to bridge the gap between academic algorithms and practical mobile visual tracking applications.
		
	To address the above mentioned issues, we propose a Distilled Siamese Tracker (DST) framework build upon what we call the Teacher-Students Knowledge Distillation (TSsKD) model, which is designed for learning a small, fast and accurate Siamese tracker through Knowledge Distillation (KD) techniques.
	TSsKD essentially explores a \textit{one teacher vs. multiple students} learning mechanism inspired by the common teaching and learning methods in schools, \textit{i.e.}, multiple students learn from one teacher, as well as help each other to facilitate learning.
	In particular, TSsKD models two kinds of KD styles. First, knowledge transfer from teacher to student is achieved by a tracking-specific distillation strategy. Second, mutual learning between students is conducted in a student-student knowledge sharing manner.
	
	More specifically, to inspire more efficient and tracking-specific KD within the same domain (without additional data or labels), the teacher-student knowledge transfer is equipped with a set of carefully designed losses, \textit{i.e.}, a Siamese target response loss, and a new distillation loss including the teacher soft and adaptive hard loss.
	The former, incorporated with a Siamese structure, is employed to learn the middle-level semantic cues, whereas the latter allows the students to mimic the high-level semantic information of the teacher and ground-truth while reducing over-fitting.
	To further enhance the performance of the students, we introduce a knowledge sharing strategy with a conditional sharing loss that encourages reliable knowledge to be shared between students. This provides extra guidance that enables small-sized trackers (the ``dim'' students) to establish a more comprehensive understanding of the tracking knowledge and thus achieve higher accuracy.
	
To evaluate the proposed algorithm, we apply it to three types of Siamese-based trackers: SiamFC~\cite{bertinetto2016fully}, SiamRPN~\cite{li2018high} and SiamRPN++~\cite{li2019siamrpn++}.
Extensive experimental evaluations on multiple tracking benchmarks demonstrate the generality and impressive performance of the proposed framework.
The distilled trackers for SiamFC and SiamRPN achieve compression rates of more than 13$\times$\!~--\!~18$\times$ and a speedup of nearly 2$\times$\!~--\!~3$\times$, respectively, while maintaining {the similar tracking accuracy.}
{As shown in Fig.~\ref{fig:example}, the distilled SiamRPN also obtains state-of-the-art performance, while operating at a high speed of 265 frames per second (FPS). In addition, we also improve the performance of the complex SiamRPN++ tracker on several existing backbones to further demonstrate the effectiveness of the proposed knowledge distillation methods.}

	To summarize, we propose an approach with the following contributions.
	\begin{itemize}
		\item A novel \textbf{Distilled Siamese Tracker (DST)} framework is proposed to compress deep Siamese-based trackers for high-performance visual tracking. To the best of our knowledge, we are the first to introduce knowledge distillation for {Siamese trackers}.
		\item Our framework is based on a \textbf{Teacher-Students Knowledge Distillation (TSsKD)} model proposed for better knowledge distillation via simulating the popular teaching mechanism among one teacher and multiple students, including teacher-student knowledge transfer and student-student knowledge sharing.
		\item In the teacher-student knowledge transfer model, we propose a novel transfer learning approach to capture the knowledge in teacher networks. Specifically, we design a \textbf{Siamese Target Response (STR)} learning algorithm to tightly couple the Siamese structure in order to effectively extract the tracking-specific knowledge and improve the performance.	
		\item \textbf{A conditional sharing loss} is proposed to transfer reliable knowledge and reduce the propagation of inaccuracy information during the knowledge sharing process. This helps address challenging tracking attributes, especially \textit{object deformation} and \textit{background clutter}.		
	\end{itemize}
	
	\section{Related Work}
	\subsection{Visual Tracking and Siamese Trackers}
In recent years, Discriminative Correlation Filter (DCF) based object tracking approaches have shown to achieve promising results. The initial work of~\cite{bolme2010visual} exploited the properties of circular correlation for training a regressor in a sliding-window fashion. Initially, DCF based trackers incorporated multi-channel hand-crafted features, such as HOG and Color Names~\cite{naresh2013correlation, danelljan2014adaptive, kiani2013multi}. Other than improved feature representations, the DCF tracking framework has been improved by integrating scale estimation~\cite{danelljan2014accurate, li2014scale} , non-linear kernels~\cite{henriques2012exploiting, henriques2014high}, and long-term memory~\cite{ma2015long}. In \cite{danelljan2016beyond}, Danelljan \textit{et al.} improved DCF via continuous operators and deep features, achieving high performance on multiple tracking benchmarks. Afterwards, several high-performance DCF trackers~\cite{kiani2017learning, dai2019visual, kart2019object, sun2019roi} have further shown to improve the tracking performance. Deep learning was first introduced to tracking in~\cite{wang2013learning}, {leading to further exploration in \cite{han2021fuse,nam2015learning,bertinetto2016fully,bertinetto2016learning,tao2016siamese,qi2018hedging,yang2020release}.}
{For example, Qi \textit{et al.}~\cite{qi2018hedging} proposed a hedge method to combine deep features from different CNN layers to better distinguish target objects.
Recently, Yang \textit{et al.}~\cite{yang2020release} improved the tracking accuracy of the deep model via online training.}

    Among deep object tracking methods, Siamese trackers \cite{li2018high,valmadre2017end,dong2018triplet,dong2018hyperparameter,shen2020hierarchical} have shown a good trade-off between speed and performance, thereby gradually becoming a mainstream tracking paradigm recently. In these trackers, the Siamese network is usually first used as a matching function in the offline phase and then applied to seek the instance in a given frame that is most similar to the initial object exemplar provided in the first frame.
    Tao \textit{et al.}~\cite{tao2016siamese} utilized a Siamese network with convolutional and fully-connected layers for training. Their approach achieved favorable accuracy, while maintaining a low speed of 2 FPS. To improve the tracking speed, Bertinetto \textit{et al.}~\cite{bertinetto2016fully} proposed ``SiamFC'', in which an end-to-end Siamese network only with five fully-convolutional layers is applied for offline training.
    The similarity between the instances in a given frame and the initial exemplar are then compared during online tracking, without complex fine-tuning strategies.
    There has been a surge of interest around SiamFC due to its high speed (nearly 86 FPS on a GPU), favorable accuracy, and simple mechanism for online tracking. Various improved methods have been proposed~\cite{li2018high,zhu2018distractor,li2019siamrpn++,fan2019siamese,zhang2019deeper,Dong2019Quadruplet,qi2020siamese,Liang2020local}. For instance, 
    Li \textit{et al.}~\cite{li2018high} proposed the SiamRPN tracker by combining the Siamese network and RPN~\cite{ren2015faster}.
    This model directly obtains the location and scale of objects by regression, avoiding the multiple forward computations required for scale estimation in common Siamese trackers. Thus, it can run at 160 FPS with a better tracking accuracy.
    In the recent VOT-2018 challenge~\cite{Kristan2018a}, a variant of SiamRPN with a larger model size won the real-time tracking task. Li \textit{et al.}~\cite{li2019siamrpn++} used a deeper backbone and a new correlation method to  propose the high-performance SiamRPN++.
    {Recently, Qi \textit{et al.}~\cite{qi2020siamese} propose a Siamese local and global network to achieve high performance on face tracking.}
	
	\subsection{Model Compression and Knowledge Distillation}
	In model compression, the aim is to squeeze a neural network to make it more efficient. In this area, pruning, quantization, and knowledge distillation are the main solutions. For pruning and quantization, compression is achieved by removing or simplifying the operations in neural networks. Srinivas \textit{et al.}~\cite{srinivas2015data} proposed to directly remove redundant neurons with low activation in a data-free way. Han~\textit{et al.}~\cite{han2015deep} proposed to remove the redundant connection and quantize the weights using Huffman coding. Gupta \textit{et al.}~\cite{gupta2015deep} used a 16-bit fixed-point representation to reduce the number of float point operations and the memory of neural networks. Further, many studies have directly trained light-weight convolutional neural networks (CNNs) with binary weights~\cite{courbariaux2015binaryconnect, courbariaux2016binarized, rastegari2016xnor}. The disadvantages of both solutions are: pruning requires many iterations before covergence and the performance is sensitive to the manually-set pruning threshold, while the performance of quantization methods is usually poor on CNNs, particularly large models.

    The idea of knowledge distillation (KD) is to improve a student network's performance by transferring knowledge from a stronger teacher network. As an earlier work, Bucilua \textit{et al.}~\cite{bucilu2006model} compressed key information from an ensemble of networks into a single neural network. More recently, Ba \textit{et al.}~\cite{ba2014deep} introduced an approach to improve the performance of shallow neural networks by mimicking deep networks in training. In another recent work, Romero \textit{et al.}~\cite{romero2014fitnets} approximated the mappings between student and teacher hidden layers to compress networks by training the relatively narrower students with linear projection layers. Subsequently, Hinton \textit{et al.}~\cite{hinton2015distilling} proposed to extract dark knowledge from a teacher network by matching the full soft distribution between the student and teacher networks during training. Following this work, KD has attracted significant interest with a variety of methods~\cite{sadowski2015deep,urban2016deep,zagoruyko2016paying,czarnecki2017sobolev,chen2017learning,zhang2018deep,furlanello2018born,kim2018paraphrasing}.
    In most existing works concerned with KD, the architecture of the student network is usually manually designed. Net-to-Net (N2N)~\cite{ashok2017n2n} is an exception to this, which focuses on automatically generating an optimally reduced architecture for KD using deep reinforcement learning.
    In the context of different computer vision applications, Chen \textit{et al.}~\cite{chen2017learning} and Liu \textit{et al.}~\cite{liu2019structured} used KD to speed up detection and segmentation networks, respectively.
    {Further, Wang~\cite{wang2020real} proposed to learn a more compact backbone for faster feature extraction in correlation filter trackers. Different to these approaches, we do not merely learn a backbone but directly obtain more efficient Siamese trackers.
	}
	
	\section{Revisiting SiamFC and SiamRPN}
	Here, we revisit the structures and training losses of both SiamFC~\cite{bertinetto2016fully} and SiamRPN~\cite{li2018high}, since they are adopted as base trackers in our distilled tracking framework.
    SiamFC adopts a two-stream fully convolutional network architecture, which takes target patches (denoted as $z$) and current search regions (denoted as $x$) as inputs. After applying a no-padding feature extraction network $\varphi$ modified from AlexNet \cite{krizhevsky2012imagenet}, a cross-correlation operation $\star$ is conducted on the two extracted feature maps:
	\begin{equation}
	\label{eq:corr-cls}
	S = \varphi(x) \star \varphi(z).
	\end{equation}
	
	The location of the target in the current frame is then inferred according to the peak value on the correlation response map $S$. The logistic loss, \textit{i.e.}, a topical binary classification loss, is used to train SiamFC:
	\begin{equation}
	\label{eq:corr-cls}
	L^{\text{FC}}(x,z,y) = \frac{1}{|S|}\sum\nolimits_{u\in S} {log(1 + e^{(-y[u] S[u])})},
	\end{equation}
	where $S[u]$ is a real-valued score in the response map $S$ and $y[u]\!\in\!\{+1, -1\}$ is a ground-truth label.
	
	SiamRPN extends SiamFC and has the same feature extraction sub-network along with an additional RPN~\cite{ren2015faster}. The final outputs are foreground-background classification score maps and regression vectors of pre-defined anchors.
	By applying a single convolution and the cross-correlation operations $\star$ from RPN on the two feature maps, the outputs are obtained as:
	\begin{equation}
	\label{eq:corr-cls}
	S_{w\times h \times 2k}^{\text{cls}} = \text{conv1}_{\text{cls}}[\varphi(x)]\star \text{conv2}_{\text{cls}}[\varphi(z)],~
	\end{equation}
	\begin{equation}
	\label{eq:corr-reg}
	S_{w\times h \times 4k}^{\text{reg}} = \text{conv1}_{\text{reg}}[\varphi(x)]\star \text{conv2}_{\text{reg}}[\varphi(z)],
	\end{equation}
	where $k$ is the pre-defined anchor number. The template feature maps $\text{conv2}_{\text{cls}}[\varphi(z)]$ and $\text{conv2}_{\text{reg}}[\varphi(z)]$ are then used as kernels in the cross-correlation operation to obtain the final classification and regression outputs, with size $w\times h$.
	
	Training is conducted by optimizing the multi-task loss:
	\begin{equation}
	L^{\text{RPN}} = L^{\text{RPN}}_{\text{cls}}(S_{w\times h \times 2k}^{\text{cls}},G_{\text{cls}}) + L^{\text{RPN}}_{\text{reg}}(S_{w\times h \times 4k}^{\text{reg}},G_{\text{reg}}),
	\end{equation}
	where $G_{\text{cls}}$ and $G_{\text{reg}}$ are the ground-truths of the {bouding box classification and regression outputs}, $L^{\text{RPN}}_{\text{cls}}$ is a cross-entropy loss for classification, and $L^{\text{RPN}}_{\text{reg}}$ is a smooth $L_1$ loss with normalized coordinates for regression. 
	
	\begin{figure*}
		\centering
		\includegraphics[width = 1 \textwidth]{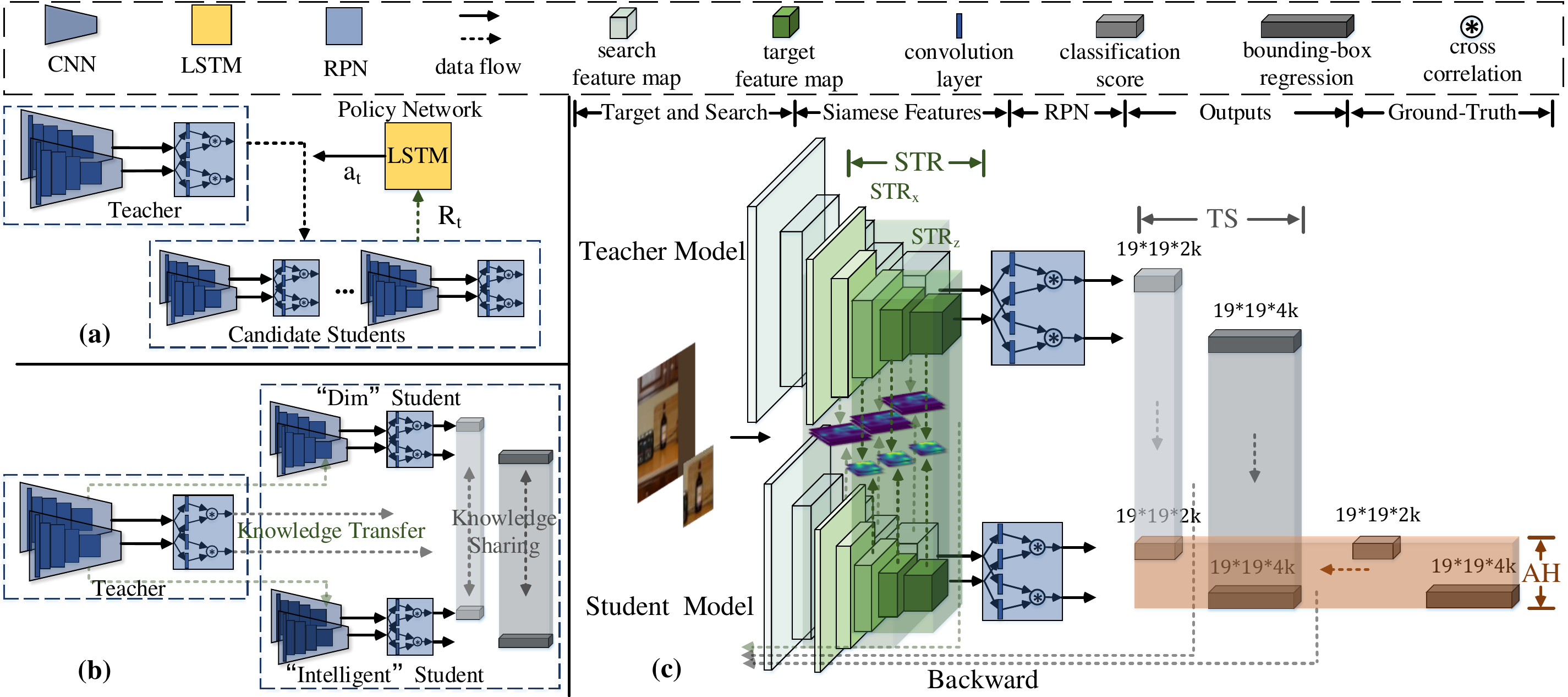}
		\caption{
			\textbf{Illustration of the proposed Distilled Siamese Tracker (DST) framework.} (a) ``Dim'' student selection via DRL: at each step $t$, a policy network guides the generation of candidate students via action $a_t$ and then updates them according to reward $R_t$. (b) Simplified schematization of our teacher-students knowledge distillation (TSsKD) model, where the teacher transfers knowledge to students, while students share knowledge with each other. (c) Detailed flow chart of teacher-student knowledge transfer with STR, TS and AH loss.
		}
		\label{fig:compression_flow}
	\end{figure*}
	
	\section{Distilled Siamese Tracker}	
	In this section, we present the proposed Distilled Siamese Tracker (DST) framework for high-performance tracking. As shown in Fig.~\ref{fig:compression_flow}, the proposed framework consists of two essential stages. First, in \S\ref{sec:selection}, for a given teacher network, such as SiamRPN, we obtain a ``dim'' student with a reduced architecture via Deep Reinforcement Learning (DRL). Then, the ``dim'' student is further simultaneously trained with an ``intelligent'' student via the proposed distillation model, facilitated by a teacher-students learning mechanism (see \S\ref{sec:KD and AT}).
	
	\subsection{``Dim'' Student Selection}
	\label{sec:selection}
	Inspired by N2N~\cite{ashok2017n2n} originally introduced for compressing classification networks, we transfer the form of selecting a student tracker with a reduced network architecture to learn an agent with an optimal compression strategy (policy) by DRL.
	{We introduce several new techniques to adapt the standard N2N classification method for Siamese trackers. This includes carefully designed variables in reinforcement learning and a practical tracking performance evaluation strategy for DRL.}

	{\noindent\textbf{Variables in Reinforcement Learning:}}
	In our task, the agent for selecting a small and reasonable tracker is learned from a sequential decision-making process by policy gradient DRL. The whole decision process can be modeled as a Markov Decision Process (MDP), which is defined as the tuple $\mathcal{M}\!=\! (\mathcal{S},\mathcal{A},\mathbb{T},R,\gamma)$. 
	
	The state space $\mathcal{S}$ is a set of all possible reduced network architectures derived from the teacher network.
	$\mathcal{A}$ is the set of all actions to transform one network into another compressed one.
	Here, we use layer shrinkage~\cite{ashok2017n2n} actions $a_t\!\in\![0.1, 0.2, \cdots, 1]$ by changing the configurations of each layer, such as kernel size, padding, and number of output filters.
	{To ensure that the size of the two networks' (teacher and student) feature maps are consistent, we only change the channel numbers of each layer. Moreover, we add a constraint on the RPN part of the SiamRPN sub-networks to force the classification and regression branches to perform the same action. This ensures that the final output channels of the networks remain unchanged after the compression operation.}
	$\mathbb{T}\!:\!\mathcal{S}\!\times\!\mathcal{A}\!\rightarrow \!\mathcal{S}$ is the state transition function, and
	$\gamma$ is the discount factor in MDP. To maintain an equal contribution for each reward, we set $\gamma$ to 1. {$R$ is the reward function, which is designed to achieve a trade-off between tracking accuracy and compression rate. The formulation is as follows: }
	\begin{equation}
	\label{eq:reward}
	R = C(2-C)\cdot \frac{\text{acc}_s}{\text{acc}_t},
	\end{equation}
	where $C\!=\!1\!-\!\frac{S_s}{S_t}$ is the relative compression rate of a student network with size $S_s$ compared to a teacher with size $S_t$, and $\text{acc}_s$ and $\text{acc}_t$ are the validation accuracy of the student and teacher networks.
	
	\noindent\textbf{Practical Tracking Evaluation:}
	To obtain an accurate evaluation of the networks' performance, we propose a new tracking performance evaluation strategy. First, we select a fixed number of images from each class of a whole dataset to form a small dataset that includes a training subset and validation subset. After tuning on the training subset, the student networks are evaluated on the validation subset.	Here, we define a new accuracy metric for tracking by selecting the top-$N$ proposals with the highest confidence and calculating their overlaps with the ground-truth boxes for $M$ image pairs from the validation subset:
	\begin{equation}
	\text{acc} = \sum\nolimits_{i=1}^{M}\sum\nolimits_{j=1}^{N}o(g_i, p_{ij}),
	\end{equation}
	where $p_{ij} (j\!\in\![1,2,\cdots,N])$ denotes the $j$-th proposal of the $i$-th image pair, $g_i$ is the corresponding ground-truth and $o$ is the overlap function.
	At each step, the policy network outputs $N_a$ actions and the reward is defined as the average reward of generated students:
	\begin{equation}
	R_t = \frac{1}{N_a} \sum\nolimits_{i=1}^{N_a}R_{t_i}.
	\end{equation}
	
	\noindent\textbf{Training Solution.}
	Given a policy network $\theta$ and the predefined MDP, we use a policy gradient network learning to compress Siamese trackers and a policy gradient algorithm to optimize the network step by step. With the parameters of the policy network denoted as $\theta$, our objective function is the expected reward over all the action sequences
	$a_{1:T}$:
	\begin{equation}
	J(\theta) = E_{a_{1:T}\sim P_{\theta}}(\mathbf{R}).
	\end{equation}
	
	We use REINFORCE \cite{williams1992simple} to calculate the gradient of our policy network. Given the hidden state $h_t$, the gradient is formulated as:
	\begin{equation}
	\label{eq_lp_re}
	\begin{aligned}
	\nabla _{\theta}J(\theta) & = \nabla _{\theta}E_{a_{1:T}\sim P_{\theta}}(\mathbf{R})\\
	& = \sum\nolimits_{t=1}^{T}E_{a_{1:T}\sim P_{\theta}}[\nabla _{\theta}\log P_{\theta}(a_t|a_{1:(t-1)})R_t]\\
	& \approx  \sum\nolimits_{t=1}^{T}[\nabla _{\theta}log P_{\theta}(a_t|h_t)\frac{1}{N_a} \sum\nolimits_{i=1}^{N_a} R_{t_i}],
	\end{aligned}
	\end{equation}
	where $P_{\theta}(a_t|h_t)$ is the probability of actions controlled by the current policy network with hidden state $h_t$. $R_{t_i}$ is the reward of the current $k$-th student model at step $t$. Furthermore, in order to reduce the high variance of estimated gradients, a state-independent baseline $b$ is introduced:
	\begin{equation}
	b = \frac{1}{N_a \cdot T}  \sum\nolimits_{t=1}^{T} \sum\nolimits_{i=1}^{N_a} R_{t_i}.
	\end{equation}
	
	It denotes an exponential moving average of previous rewards. Finally, our policy gradient is calculated as:
	\begin{equation}
	\nabla _{\theta}J(\theta)
	\approx  \sum\nolimits_{t=1}^{T}[\nabla _{\theta}log P_{\theta}(a_t|h_t)(\frac{1}{N_a}\sum\nolimits_{i=1}^{N_a}R_{t_i}-b)].
	\end{equation}
	We use the gradient to optimize the policy, obtaining a final policy $\pi_\theta\!:\!\mathcal{S}\!\rightarrow\!\mathcal{A}$ and reduced student network. All the training processes in this section are based on the small dataset selected from the whole dataset, considering the time cost of training all students.
	
	{Notice that our selection process is a simple network architecture search (NAS) method, where the search space includes the networks with the same layers but fewer convolution channels. Our method aims to search a student network with relatively good performance. A more advanced NAS method may get a better network structure, and it is one effective technique for practical engineering applications. However, this is not our research focus. Thus, we only use a simple method as a baseline.}
	
	\subsection{Teacher-Students Knowledge Distillation}
	\label{sec:KD and AT}
	After the network selection, we obtain a ``dim'' student network with poor comprehension {due to the small model size}. To pursue more intensive knowledge distillation and favorable tracking performance, we propose a Teacher-Students Knowledge Distillation (TSsKD) model. It encourages teacher-student knowledge transfer as well as mutual learning between students, which serves as more flexible and appropriate guidance. In \S\ref{sec:single TS}, we elaborate the teacher-student knowledge transfer (distillation) model. Then, in \S\ref{sec:KS}, we describe the student-student knowledge sharing strategy.  	
	
	\subsubsection{{Teacher-Student Knowledge Transfer}}
	\label{sec:single TS}	
{In the teacher-student knowledge transfer model, we introduce a transfer learning approach to capture the knowledge in teacher networks. It contains two components: the Siamese Target Response (STR) learning and distillation loss. The former is used for transferring the middle-level feature maps without background disturbance to the student from the teacher. This provides clean critical middle-level semantic hints to the student.
The latter includes two sub-loss: the Teacher Soft (TS) loss and Adaptive Hard (AH) loss, which allows the student to mimic the high-level outputs of the teacher network, such as the logits~\cite{hinton2015distilling} in the classification model.
This loss can be viewed as a variant of KD methods~\cite{hinton2015distilling,chen2017learning}, which are used to extract dark knowledge from teacher networks.
    Our transfer learning approach includes both classification and regression and can be incorporated into other networks by removing the corresponding part.}
	
	\noindent\textbf{Siamese Target Response (STR) Learning.}
	In order to lead a tracker to concentrate on the same target as a teacher, we propose a background-suppression Siamese Target Response (STR) learning method in our framework. Based on the assumption that the activation of a hidden neuron can indicate its importance for a specific input, we transfer the semantic interest of a teacher onto a student by forcing it to mimic the teacher's response map. We gather the feature responses of different channels into a response map by a mapping function $F\!:\!\mathbb{R}_{C \times H \times W}\!\to\!\mathbb{R}_{H \times W}$, which outputs a 2D response map from the 3D feature maps provided. We formulate this as:
	\begin{equation}
	\label{eq:att-map}
	F(U) = \sum\nolimits_{i=1}^{C}|U_i|,
	\end{equation}
	where $U_i \in \mathbb{R}_{H \times W}$ is the $i$th channel of a spatial feature map, and $|\cdot|$ represents the absolute values of a matrix. In this way, we squeeze the responses of different channels into a single response map.
	
	Siamese trackers have two weight-sharing branches with different inputs: a target patch and a larger search region.
	{Different from the detection~\cite{chen2017learning} requiring multiple responses on objects in the search region, our tracking task only needs one ideal high target response in both search region and template patch, respectively.}
	To learn the target responses of both branches, we combine their learning process. Since we found that the surrounding noise in the search region's response will disturb the response learning of the other branch due to the existence of distractors, we set a weight on the search region's feature maps. The following multi-layer response learning loss is defined:
	\begin{equation}
	\label{eq:AT-loss}
	L^{\text{STR}} = L^{\text{STR}}_{x} + L^{\text{STR}}_{z},
	\end{equation}
	\begin{equation}
	\label{eq:AT-loss-x}
	L^{\text{STR}}_{x} = \sum\nolimits_{j \in \tau}^{} \| F(W_S^j Q_{S_x}^j) - F(W_T^j Q_{T_x}^j) \|_2,
	\end{equation}
	\begin{equation}
	\label{eq:AT-loss-z}
	L^{\text{STR}}_{z} = \sum\nolimits_{j \in \tau}^{} \| F(Q_{S_z}^j) - F(Q_{T_z}^j) \|_2,~~~~~~~~
	\end{equation}
	where $\tau$ is the set of layers' indices conduct. $Q_{T_x}^j$ and $Q_{T_z}^j$ denote the teacher's feature map of layer $j$ in the search and target branch, respectively. $W_T^j\!=\!Q_{T_x}^j\!\star\!Q_{T_z}^j$ is the weight on the teacher's $j$th feature map. {The students' variables, such as $W_{S}^j$, are defined in the same way.}
	
\begin{figure}
		\centering
		\small
		\includegraphics[width = .5 \textwidth]{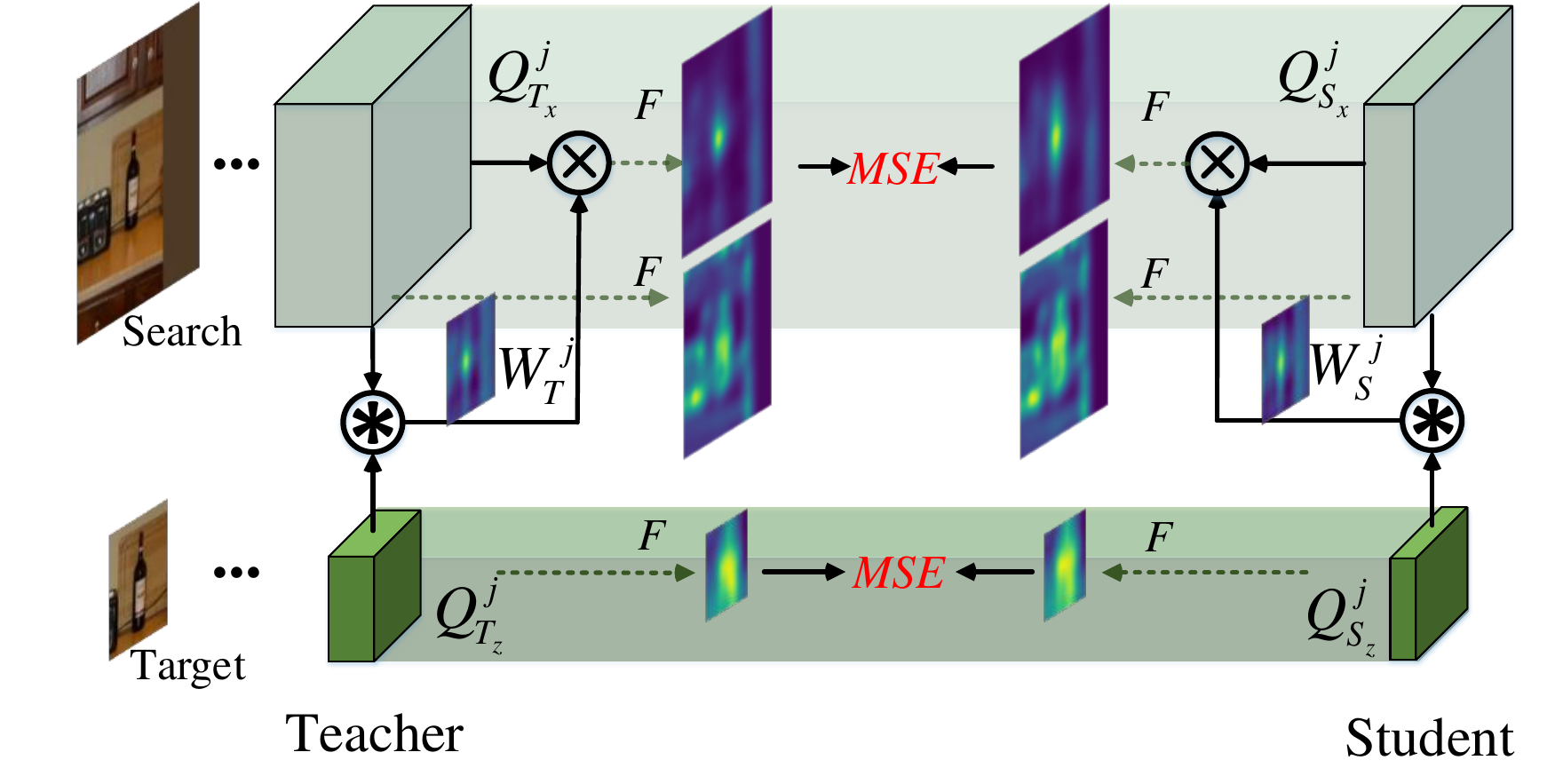}
		\caption{\textbf{Illustration of our Siamese Target Response (STR) learning.} Take one layer as an example. For the target branch, feature maps are directly transformed into 2D activation maps. For the search branch, weights ($W_T^j$ and $W_S^j$) are calculated by conducting a cross-correlation operation on the two branches' feature maps and then multiplying the result by the search feature map.}
		\label{fig:STRLoss_example}
	\end{figure}

	By introducing this weight, which rearranges the importance of different areas in the search region according to their similarities with the target, the response activation is concentrated on the target. This keeps the response maps from the two branches consistent and enhances the response learning. An example of our multi-layer Siamese target response learning is shown in Fig.~\ref{fig:STRLoss_example}. The comparison of response maps with and without weights shows that the surrounding noise is suppressed effectively. {It is worth mentioning that our STR method is the first attempt to do feature learning on Siamese networks.}
	
	\noindent\textbf{Distillation Loss.}
	{Our distillation loss contains two critical components: the teacher soft loss, which is used to imitate the teacher, and the adaptive hard loss which is designed to integrate the ground-truth.}
	
	\textit{1) Teacher Soft (TS) Loss.}
	We set $C_s$ and $B_s$ as the student's classification and bounding box regression outputs, respectively. {$C_t$ and $B_t$ represent the teachers' variables.} In order to incorporate the dark knowledge~\cite{hinton2015distilling} that regularizes students by placing emphasis on the relationships learned by the teacher network across all the outputs, we need to `soften' the output of classification. We set $P_t\!=\!\text{softmax}(C_t/\text{temp})$, where $\text{temp}$ is a temperature parameter used to obtain a soft distribution~\cite{hinton2015distilling}. Similarly, $P_s\!=\!\text{softmax}(C_s/\text{temp})$. Then, we formulate the TS loss as follows:
	\begin{equation}
	\label{eq:TS-loss}
	L^{\text{TS}} = L_{\text{cls}}^{\!\text{TS}}(P_s, P_t) + L_{\text{reg}}^{\!\text{TS}}({B_s, B_t}),
	\end{equation}
	where $ L_{\text{cls}}^{\text{TS}}\!=\!\text{KL}(P_s, P_t)$ is a Kullback Leibler (KL) divergence loss on the soft outputs of the teacher and student. $L_{\text{reg}}^{\text{TS}}$ is the original regression loss of the tracking network.
	
	\textit{2) Adaptive Hard (AH) Loss.}
	To make full use of the ground-truth $G$, we combine the outputs of the teacher network with the original hard loss of the student network. For the regression loss, we employ a \textit{teacher bounded regression loss}~\cite{chen2017learning}.
	\begin{equation}
	L_{\text{reg}}^{\text{AH}}(B_s, B_t, G_{\text{reg}})=
	\begin{cases}
	L_r(B_s,G_{\text{reg}}), & \text{if} ~~gap < m, \\
	0,	& \text{otherwise}.
	\end{cases}
	\end{equation}
	where $ gap\!=\!L_r(B_t,G_{\text{reg}})\!-\!L_r(B_s,G_{\text{reg}})$ is the gap between the student's and the teacher's loss ($L_r$ is the regression loss of the tracking network) with the ground-truth.
	$m$ is a margin. This loss aims to keep the student's regression vector close to the ground-truth when its quality is worse than the teacher. However, once it {is close to or outperforms} the teacher network, {we remove the loss for the student} to avoid over-fitting.
	{For the classification loss, there is no unbounded issue in the discrete classification task~\cite{chen2017learning}. Thus, we directly use the student's original classification loss $L_{\text{cls}}^{\text{AH}}$. Finally, our AH loss is defined as follows:}
	\begin{equation}
	\label{eq:AH-loss}
	L^{\text{AH}} = L_{\text{cls}}^{\text{AH}}(C_s, G_{\text{cls}}) + L_{\text{reg}}^{\text{AH}}(B_s, B_t, G_{\text{reg}}).
	\end{equation}
	
	{Then, incorporated with the TS loss, our distillation loss is formulated as,
	\begin{equation}
	\label{eq:distillation-loss}
	L^{\text{D}} =\eta L^{\text{TS}} +  \lambda L^{\text{AH}},
	\end{equation}
	where $\eta$ and $\lambda$ are balance parameters. It is worth mentioning that our $L^{\text{D}}$ is not a simple extension of the distillation loss for object detection (denoted as $L^{\text{D}}_{\text{O}}$) in \cite{chen2017learning}. To closely analyze the differences, we reformulate it as the summary of the soft loss and hard loss, {\it i.e.} $L^{\text{D}}_{\text{O}} = L^{\text{S}}_{\text{O}} + L^{\text{H}}_{\text{O}}$
	, where
	\begin{equation}
	\label{eq:soft-loss}
	L^{\text{S}}_{\text{O}} =(1-\mu) L_{\text{cls}}^{\!\text{TS}}(P_s, P_t),
	\end{equation}
	\begin{equation}
	\label{eq:hard-loss}
	L^{\text{H}}_{\text{O}} =\mu L_{\text{cls}}^{\text{AH}}(C_s, G_{\text{cls}}) + L_{\text{reg}}^{\text{AH}}(B_s, B_t, G_{\text{reg}}) + \nu L_r(B_s,G_{\text{reg}}).
	\end{equation}

	To focus on the major components of these losses, we ignore the impact of the balance parameters ($\mu$, $\nu$). Comparing Eq. (\ref{eq:TS-loss}) with (\ref{eq:hard-loss}), the previous soft loss $L^{\text{S}}_{\text{O}}$ lacks the teacher regression loss $L_{\text{reg}}^{\!\text{TS}}$.
	{In the detection task, the teacher may provide wrong guidance~\cite{chen2017learning}. However, in visual tracking which can be regarded as a one-shot single target detection task, the teacher's regression output is accurate enough. Furthermore, lacking the teacher regression loss indicates that the  $L^{\text{S}}_{\text{O}}$ loss can not push the student network to imitate the regression branch of the teacher as much as possible or fully mine the underlying knowledge inside the teacher.}
	Thus, we add the regression loss $L_{\text{reg}}^{\!\text{TS}}$ into the proposed distillation loss. Comparing Eq. (\ref{eq:AH-loss}) with (\ref{eq:hard-loss}), the previous hard loss $L^{\text{H}}_{\text{O}}$ contains an additional regression loss $L_r^s = L_r(B_s,G_{\text{reg}})$, which is a little redundant for two reasons. Firstly, when the student performs worse than the teacher ($gap<m$), $L_r^s$ provides the same information as $L_{\text{reg}}^{\text{AH}}$. If the student performs similar to or better than the teacher network, which means that the student has learnt enough knowledge from the ground-truth, $L_r^s$ will not offer additional information for training. In contrast, the strong supervision from $L_r^s$ easily leads to over-fitting on some samples. Therefore, we remove $L_r^s$ in our distillation loss.
	}

	\noindent\textbf{Overall Loss.}
	{By combining the above STR loss and distillation loss,} the overall loss for transferring knowledge from a teacher to a student is defined as follows:
	\begin{equation}
	\label{eq:transfer-loss}
	L^{\text{KT}} =\omega L^{\text{STR}} + \eta L^{\text{TS}} +  \lambda L^{\text{AH}}.
	\end{equation}
	
	\subsubsection{Student-Student Knowledge Sharing}
	\label{sec:KS}
	Based on our teacher-student distillation model, we propose a student-student knowledge sharing mechanism to further narrow the gap between the teacher and the ``dim'' student. As an ``intelligent'' student with a larger model size usually learns and performs better (due to its better comprehension), sharing its knowledge is likely able to inspire the ``dim'' one to develop a more in-depth understanding. On the other side, the ``dim'' student can do better in some cases and provide some useful knowledge too. {To capture the useful knowledge and reduce the ``bad'' knowledge from the ``dim'' student, we propose a conditional suppressed weighting for our knowledge sharing.}
	{It is worth mentioning that our experimental results show the proposed knowledge sharing is effective for a variety of tracking challenges, especially in case of \textit{deformation} and \textit{background clutter}. More details are presented in \S\ref{sec:siamrpn&siamfc4}.}
	
	{
	We take two students as an example and denote them as a ``dim'' student $\text{s1}$ and an ``intelligent'' student $\text{s2}$.
	For a true distribution $p(x)$ and the predicted distribution $q(x)$, the KL divergence on $N$ samples is defined as:
	\begin{equation}
	\begin{aligned}
	D_{KL}(p||q) &= \sum_{i=1}^N p(x_i)log(\frac{p(x_i)}{q(x_i)})\\
	&= - H(p(x)) + (-\sum_{i=1}^N p(x_i)log(q(x_i)))\\
	&= - H(p(x)) + L_{CE}(p, q),
	\end{aligned}
	\end{equation}
	where $H$ is the entropy and $L_{CE}$ is the cross-entropy. We can see that, unlike the traditional cross-entropy loss, the KL divergence loss contains an entropy item of the label (true distribution). The tracking classification output can be regarded as a probability distribution function (PDF). If $p(x)$ is a hard label and thus a fixed constant, the KL divergence and cross-entropy loss are equal during training. However for the knowledge sharing, $p(x)$ is a differentiable output of the other student. Compared to the cross-entropy loss, the KL divergence (also known as relative entropy) offers a more accurate supervision signal for the training and can better measure the similarity of two PDFs. Therefore, we use KL divergence loss as the classification loss for knowledge sharing.
	}
	
	For a proposal $d_i$ in a Siamese tracker, assume that the predicted probabilities of being target by $\text{s1}$ and $\text{s2}$ are $p_1(d_i)$ and $p_2(d_i)$, respectively. The predicted bounding-box regression values are $r_1(d_i)$ and $r_2(d_i)$. To improve the learning effect of $\text{s1}$, we obtain the knowledge shared from $\text{s2}$ by using its prediction as prior knowledge.
	The KL divergence is defined as:
	\begin{equation}
	\label{eq:KS-cls}
	\!\!L_{\text{cls}}^{\text{KS}}\!(\text{s1}||\text{s2})\!
	=\!\!\sum_{i=1}^N \!(p_1(d_i)log\frac{p_1(d_i)}{p_2(d_i)} \!+\! (1-p_1(d_i))log\frac{1-p_1(d_i)}{1-p_2(d_i)}).
	\end{equation}

	For regression, we use the smooth $L_1$ loss:
	\begin{equation}
	\label{eq:KS-reg}
	L_{\text{reg}}^{\text{KS}}(\text{s1}||\text{s2}) = \sum_{i=1}^N L_1(r_1(d_i)-r_2(d_i)).
	\end{equation}
	The knowledge sharing loss for $s1$ can be defined as:
	\begin{equation}
	\label{eq:KS-loss1}
	L^{\text{KS}}(\text{s1}||\text{s2}) =  L^{\text{KS}}_{\text{cls}}(\text{s1}||\text{s2}) + L^{\text{KS}}_{\text{reg}}(s1||s2).
	\end{equation}
	
	{
	However, there exists some representation inaccuracy during the sharing process, especially at the first several epochs. To filter out the reliable shared knowledge, we introduce a {conditional suppressed weighting function} $\sigma$ for $L^{\text{KS}}$:
    \begin{equation}
	\label{eq:sigma}
	\small
	\sigma(\text{s1}) =
	\begin{cases}
	f(e)& \text{if}~L^{\text{GT}}(\text{s2})-L^{\text{GT}}(t)<h, \\
	0& \text{otherwise}.
	\end{cases}
	\end{equation}
	Here, $L^{\text{GT}}(\text{s2})$, $L^{\text{GT}}(t)$ are the losses for s2 and the teacher, with ground-truth. $h$ is their gap constraint. $f(e)$ is a function that decreases geometrically with current epoch $e$. Our knowledge transfer and sharing run simultaneously, which prevents the accumulation of the representation errors. The overall knowledge distillation loss with conditional knowledge sharing is defined as:
	\begin{equation}
	\label{eq:final-loss}
	L^{\text{KD}}_{\text{s1}} =  L^{\text{KT}}_{\text{s1}} + \sigma (\text{s1}) L^{\text{KS}}(\text{s1}||\text{s2}).
	\end{equation}
	In this way, we ensure that the shared knowledge offers accurate guidance to s1 and enhances the training. For s2, the loss is similar:
	\begin{equation}
	L^{\text{KD}}_{\text{s2}} =  L^{\text{KT}}_{\text{s2}} + \beta\cdot \sigma(\text{s2}) L^{\text{KS}}(\text{s2}||\text{s1}),
	\end{equation}
	where $\sigma(\text{s2})$ is defined in the same way as $\sigma(\text{s1})$ and $\beta$ is a discount factor on account of the difference in reliability of the two students. Considering the ``dim'' student's worse performance, we set $\beta \in (0, 1)$.
	}
	
	{
	Finally, to train two students simultaneously, the loss for our TSsKD is:
	\begin{equation}
	L^{\text{KD}} =  L^{\text{KD}}_{\text{s1}} + L^{\text{KD}}_{\text{s2}}.
	\end{equation}
	}

	\section{Experiments}
	To demonstrate the effectiveness of the proposed method, we conduct experiments on SiamFC~\cite{bertinetto2016fully}), SiamRPN~\cite{li2018high} (VOT version as in ~\cite{li2019siamrpn++}) and SiamRPN++~\cite{li2019siamrpn++}. For the simple Siamese trackers (SiamRPN~\cite{li2018high} and SiamFC~\cite{bertinetto2016fully}), since there is no smaller classic handcrafted structure, we first search and then train a proper ``dim'' student via our framework. We evaluate the distilled trackers on several benchmarks and conduct an ablation study (from \S\ref{sec:siamrpn&siamfc1} to \S\ref{sec:siamrpn&siamfc4}). Furthermore, to validate our TSsKD on well-designed  handcrafted  structures, we distilled SiamRPN++ trackers with different backbones (\S\ref{sec:siamrpn&siamfc5}). {Note that all experiments, unless otherwise stated, are conducted using two students.}
	
	\subsection{Implementation Details}
	\label{sec:siamrpn&siamfc1}
	\noindent\textbf{Reinforcement Learning Setting.} In the ``dim'' student selection experiment, an LSTM is employed as the policy network.
	The size of its inputs and outputs is 5 and 11, respectively.
	A small representative dataset (about 10,000 image pairs) is created by selecting images uniformly from several classes in the whole dataset to train the corresponding tracker. The policy network is updated over 50 steps. In each step, three {compressed} networks are generated and trained from scratch for 10 epochs. We observe heuristically that this is sufficient to compare performance.
	After each step, the policy network is optimized via Adam with learning rate being 0.003.
	Both SiamRPN and SiamFC use the same settings.
	
	\begin{table}
		\centering
		\resizebox{0.5\textwidth}{!}{
			\begin{tabular}{c|c|c|c|c}
				\hline
				&$L^{\text{AH}}$&$L^{\text{TS}}$&$L^{\text{STR}}$&$L^{\text{KS}}$\\
				\hline
				SiamFC&logistic&KL&MSE&KL\\
				\hline
				SiamRPN&cross-entropy+bounded&KL+$L_1$&MSE&KL+$L_1$\\
				\hline
			\end{tabular}
		}
		\caption{Losses used in the knowledge transfer stage. MSE, $L_1$ and KL represent Mean-Square-Error loss, smooth $l_1$ loss and Kullback Leibler divergence loss, respectively}
		\label{losses}
	\end{table}

	\begin{figure}
		\small
		\centering
		\subfigure[]{
			\label{fig:n2n-siamrpn}
			\includegraphics[width = .24 \textwidth]{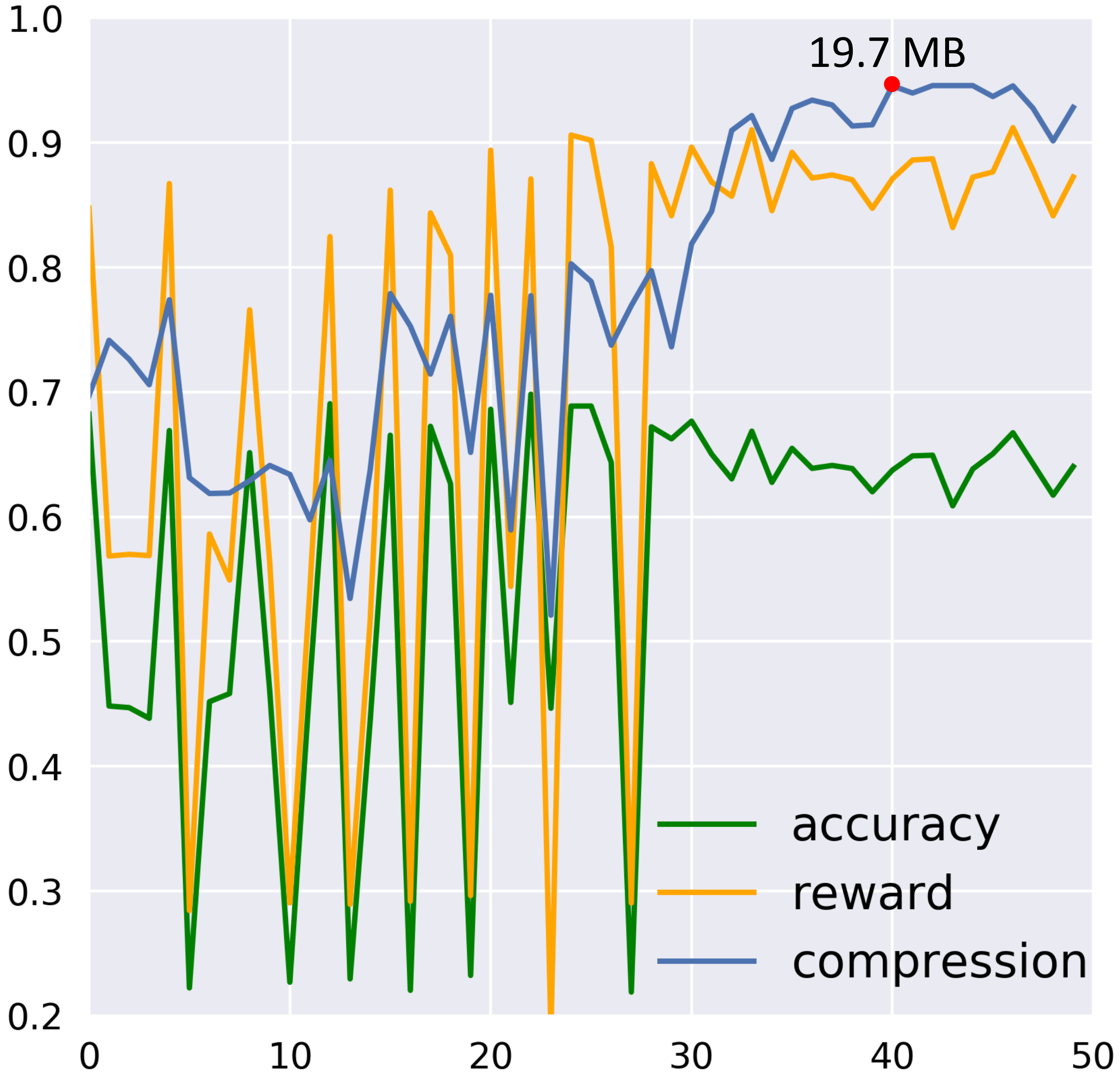}
		}
		\subfigure[]{
			\label{fig:n2n-siamfc}
			\includegraphics[width = .24 \textwidth]{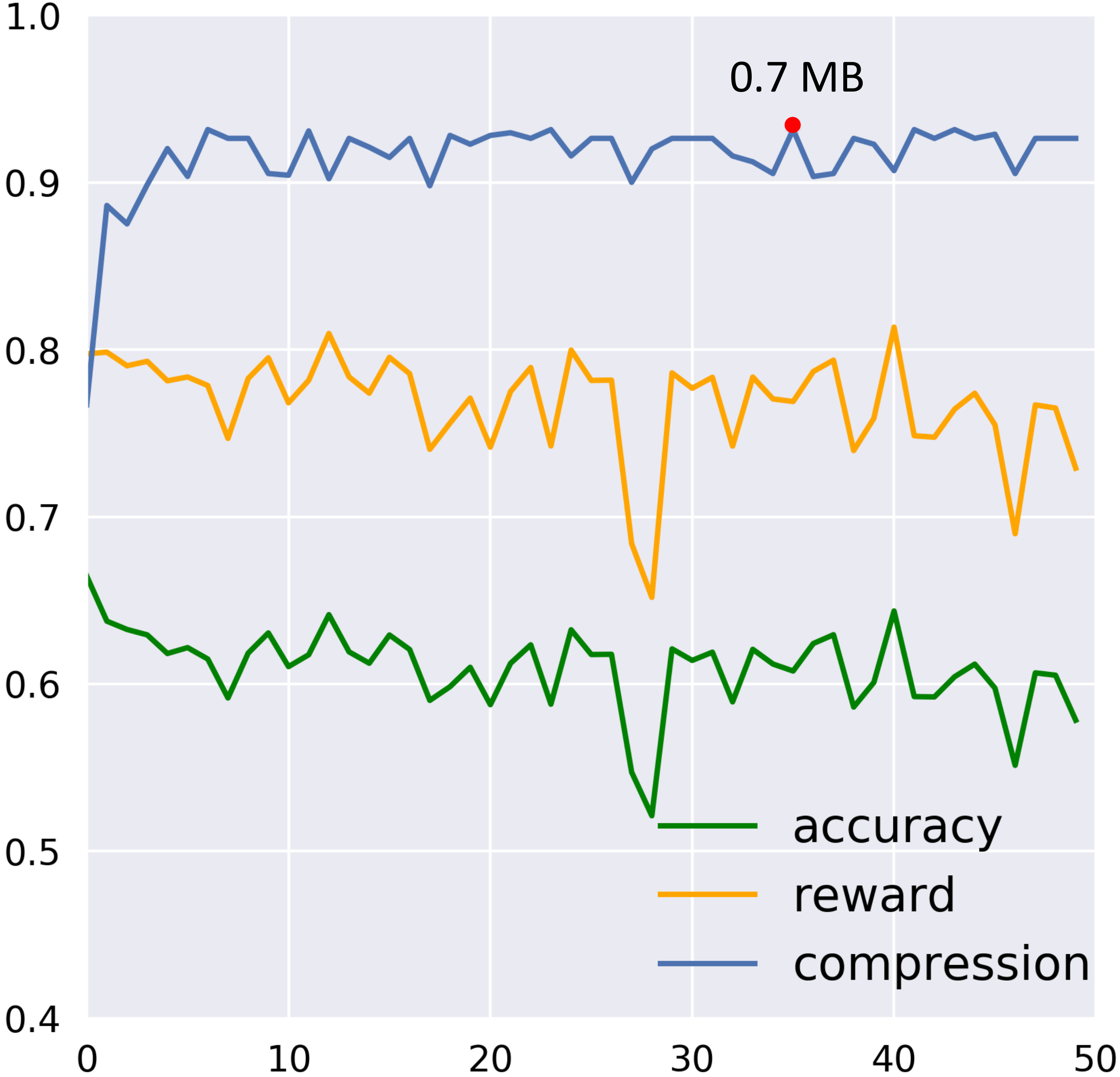}
		}
		\caption{``Dim'' student selection on (a) SiamRPN and (b) SiamFC. Reward, accuracy, compression (relative compression rate $C$ in Eq.~\ref{eq:reward}) vs. iteration.}
		\label{fig:n2n}
	\end{figure}

	\noindent\textbf{Training Datasets.}
	For SiamRPN, as with the teacher~\cite{Kristan2018a}, we pre-process four datasets: ImageNet VID~\cite{ILSVRC15}, YouTube-BoundingBoxes~\cite{real2017youtube}, COCO~\cite{lin2014microsoft} and ImageNet Detection~\cite{ILSVRC15}, to generate about two million image pairs with {127$\times$127} pixel target patches and {271$\times$271} pixel search regions. Unlike SiamRPN, our student does not need a backbone pre-trained on ImageNet and can learn good discriminative features via STR.
	To make the model robust to gray videos, 25$\%$ of the pairs are converted into grayscale during training. Moreover, a translation within 12 pixels and a resize operation varying from 0.85 to 1.15 are performed on each training sample to increase the diversity.
	Our SiamFC is trained on ImageNet VID~\cite{ILSVRC15} with 127$\times$127 pixels and 255$\times$255 pixels for the two inputs, respectively, which is consistent with SiamFC~\cite{bertinetto2016fully}.
	
	\noindent\textbf{Optimization.}
	During the teacher-students knowledge distillation, ``intelligent'' students are generated by halving the convolutional channels of {the teachers (SiamRPN and SiamFC).}
	SiamRPN's student networks are warmed up by training with the ground-truth for 10 epochs, and then trained for 50 epochs with the learning rate exponentially decreasing from $10^{-2}$ to $10^{-4}$. As with the teacher, SiamFC's student networks are trained for 30 epochs with a learning rate of $10^{-2}$. All the losses used in the experiments are reported in Table~\ref{losses}. The other hyperparameters are set to: $m=0.005$, $\omega=100$, $\eta=1$, $\lambda=0.1$, $temp=1$, $h=0.005$ and $\beta=0.5$.
	We set these weights according to the empirical settings and the scale of different loss components.

\begin{figure}
		\small
		\centering
		\subfigure[]{
			\label{fig:losses-siamrpn_train}
			\includegraphics[width = .227 \textwidth]{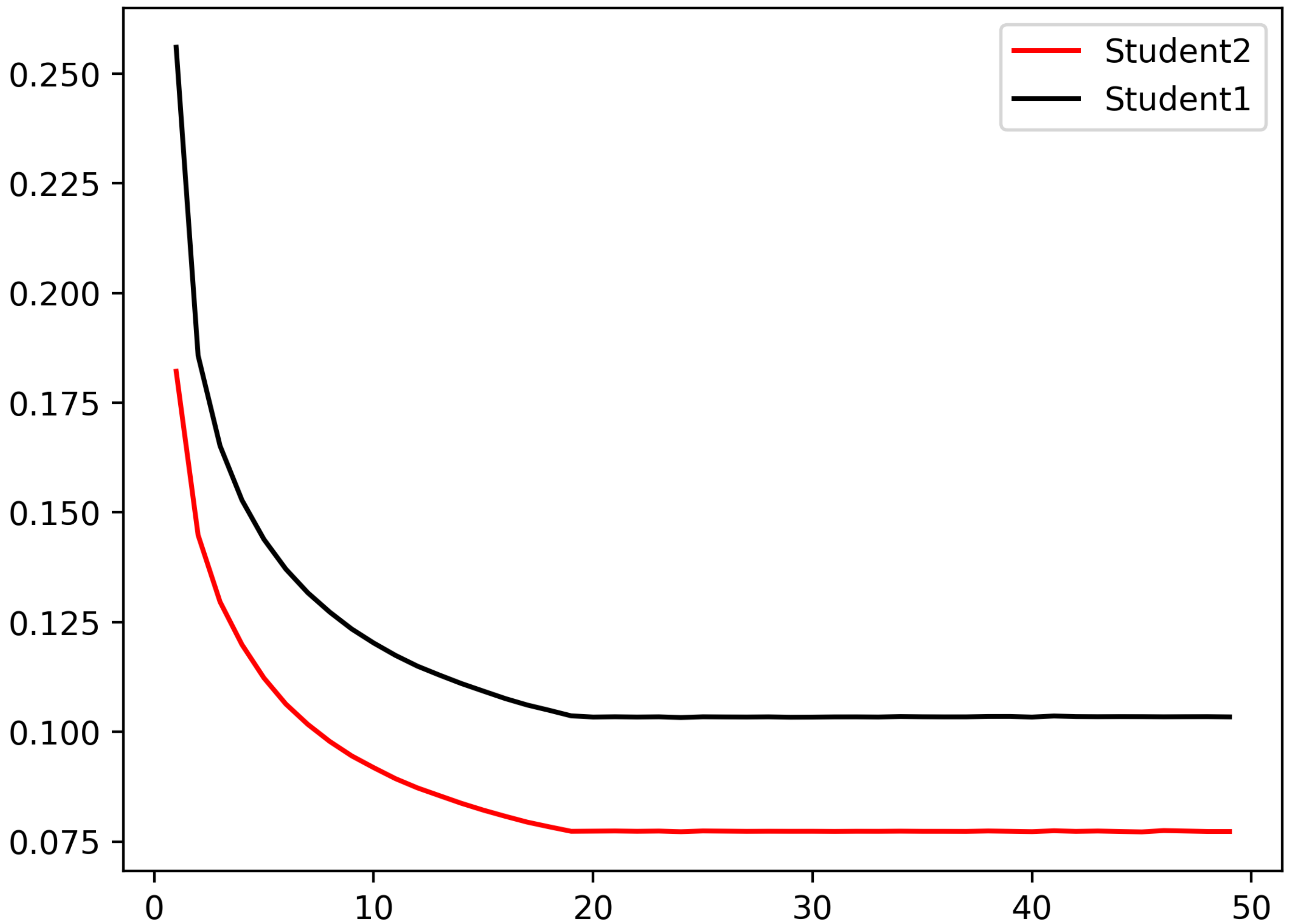}
		}
		\subfigure[]{
			\label{fig:losses-siamfc_train}
			\includegraphics[width = .227 \textwidth]{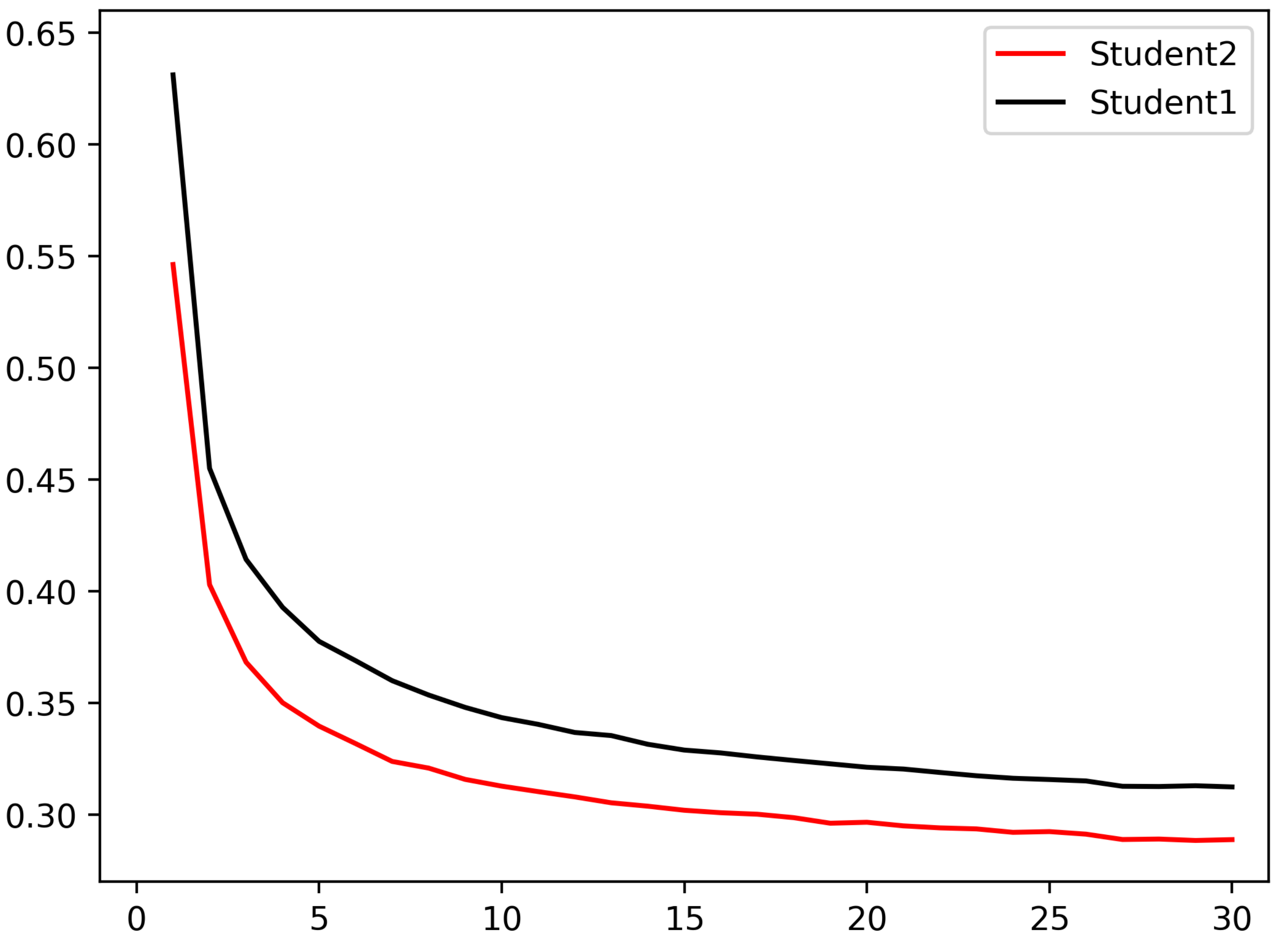}
		}
		\subfigure[]{
			\label{fig:losses-siamrpn_val}
			\includegraphics[width = .227 \textwidth]{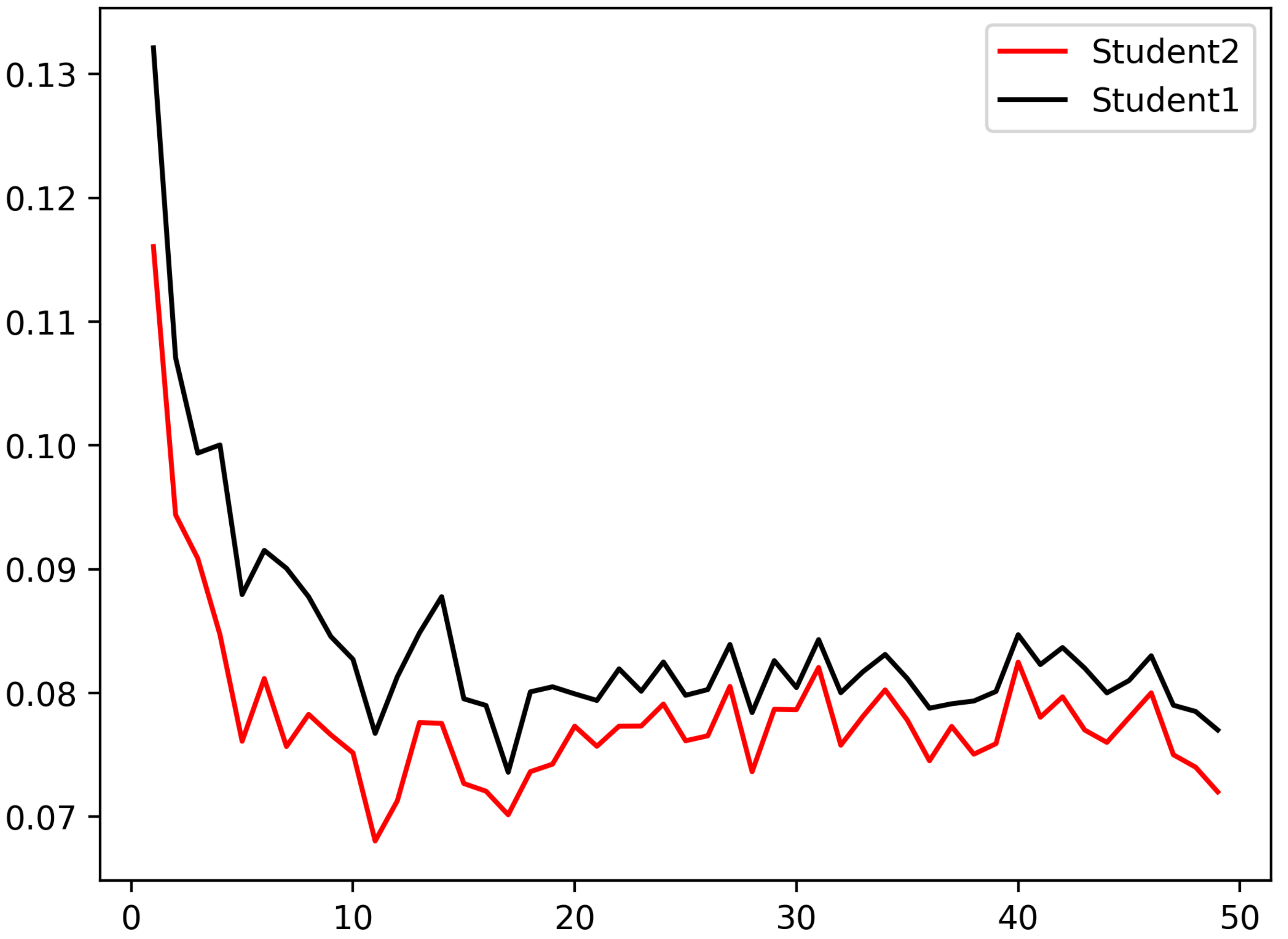}
		}
		\subfigure[]{
			\label{fig:losses-siamfc_val}
			\includegraphics[width = .227 \textwidth]{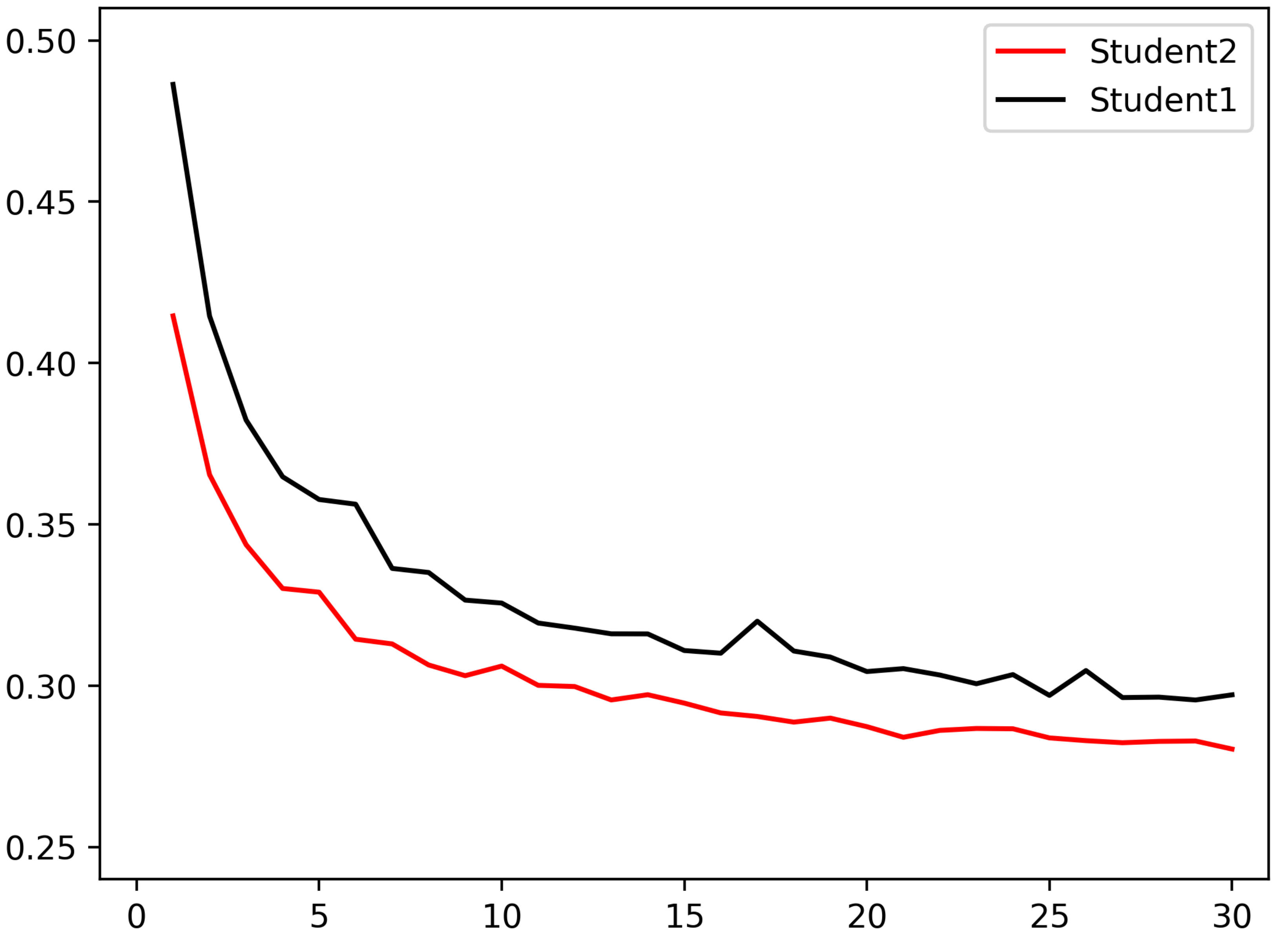}
		}
		\caption{Losses comparison, including  training loss of the (a) SiamRPN and (b) SiamFC students, and validation loss of the (c) SiamRPN and (d) SiamFC students.}
		\label{fig:losses}
	\end{figure}

	\begin{table}[!htbp]
		\small
		\centering
		\resizebox{0.5\textwidth}{!}
		{
			\begin{tabular}{c|c|c|c|c|c|c|c}
				\hline
				&conv1 &conv2 &conv3 &conv4 &conv5 &rpn\_cls1/2 &rpn\_reg1/2\\
				\hline
				\textbf{DSTfc}&(3,38) &(38,64) &(64,96) &(96,96)&(96,64)&/ &/\\
				\hline
				\textbf{DSTrpn}&(3,192) &(192,128) &(128,192) &(192,192)&(192,128)&(128,128/1280) &(128,128/2560)\\
				\hline
			\end{tabular}
		}
		\vspace*{2pt}
		\caption{Detailed convolutional structures of DSTfc and DSTrpn. The numbers denote the input and output channel numbers of the corresponding convolutional layers.}
		\label{structure}
	\end{table}
	
	\subsection{Evaluations of ``Dim'' Student Selection}
	\noindent\textbf{Training Details.} As shown in Fig.~\ref{fig:n2n-siamrpn}, the complicated architecture of SiamRPN caused several inappropriate SiamRPN-like networks to be generated in the top 30 iterations, leading to unstable accuracies and rewards.
	After five iterations, the policy network gradually converges and finally achieves a high compression rate.
	On the other side, the policy network of SiamFC converges quickly after several iterations due to its simple architecture (See Fig.~\ref{fig:n2n-siamfc}). The compression results show that our method is able to generate an optimal architecture regardless of the teacher's complexity. Finally, two reduced models of size 19.7 MB and 0.7 MB for SiamRPN (361.8 MB) and SiamFC (9.4 MB) are generated. The detailed structures of the reduced models, DSTrpn and DSTfc, are shown in Table~\ref{structure}.
	
	\noindent\textbf{Loss Comparison.}
	We also compare the losses of different student networks. As shown in Fig.~\ref{fig:losses}, the ``intelligent'' students (denoted as Student1) have a lower loss than the ``dim'' ones (denoted as Student2) throughout the whole training and validation process, and maintain a better understanding of the training dataset. They provide additional reliable knowledge to the ``dim'' students which further promotes more intensive knowledge distillation and better tracking performance.
	\subsection{Benchmark Results}
	
	\begin{figure}
		\centering
		\includegraphics[width = .235\textwidth]{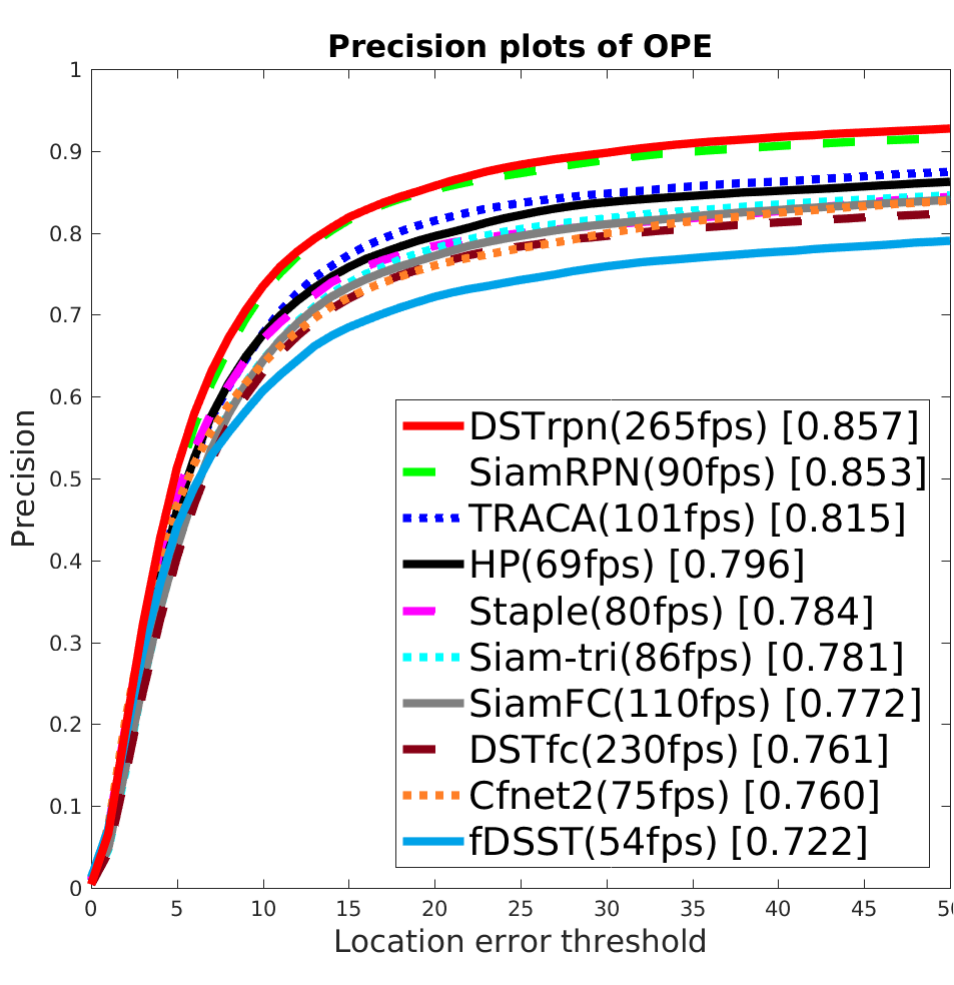}
		\includegraphics[width = .235\textwidth]{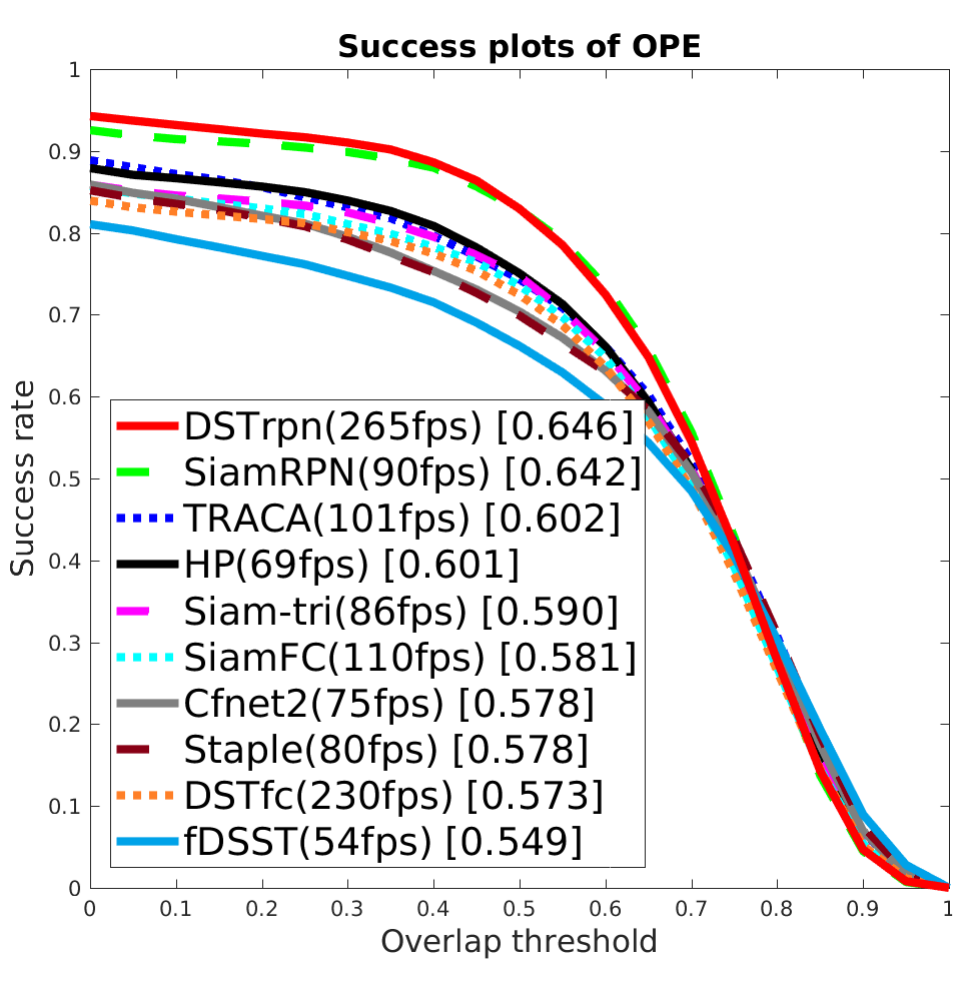}	
		\caption{Precision and success plots with AUC for OPE on the OTB-100 benchmark~\cite{wuobject2015}.}
		\label{fig:res_otb_100}
	\end{figure}
	
	\begin{figure}
		\centering
		\includegraphics[width = .24\textwidth]{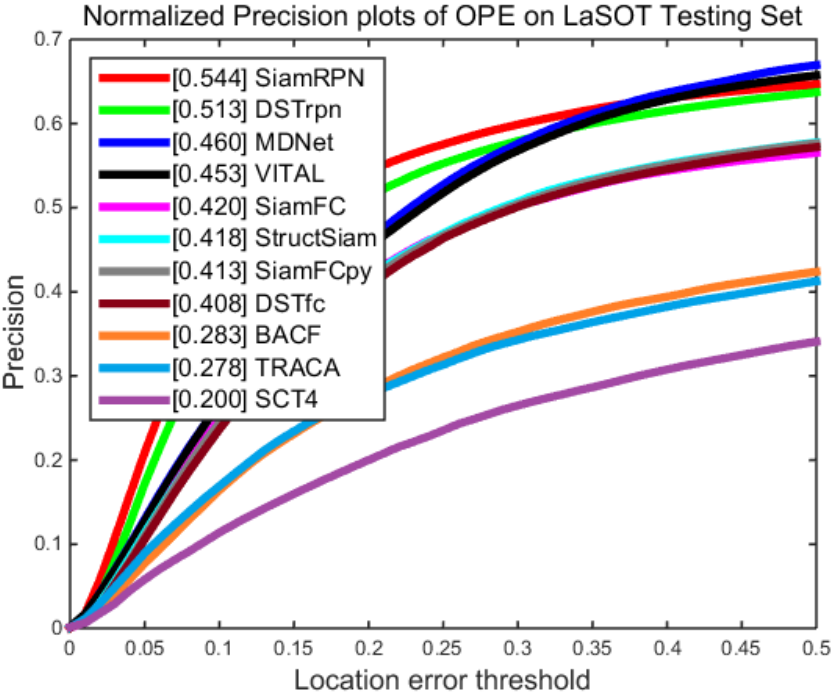}
		\centering
		\includegraphics[width = .24\textwidth]{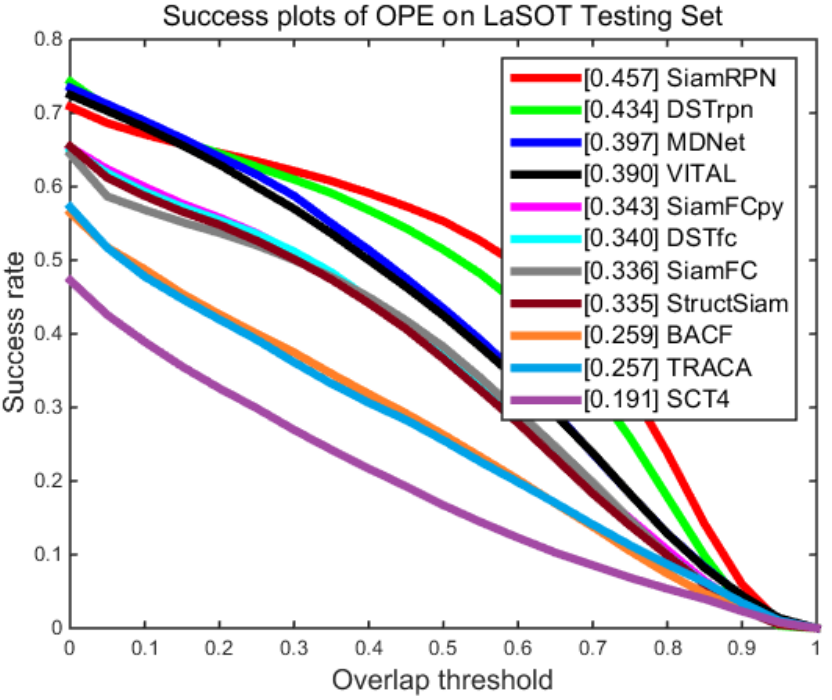}	
		\caption{Evaluation results of trackers on the LaSOT~\cite{fan2019lasot}.}
		\label{fig:res_lasot}
	\end{figure}
	
	\noindent\textbf{Results on OTB-100.}
On the OTB-100 benchmark~\cite{wuobject2015}, a conventional method for evaluating trackers is one-pass evaluation (OPE).
To measure the performance with different initializations, we use spatial robustness evaluation (SRE) and temporal robustness evaluation (TRE).
SRE uses different bounding boxes in the first frame and TRE starts at different frames for initialization.
We compare our DSTrpn (SiamRPN as teacher) and DSTfc (SiamFC as teacher) trackers with various recent fast trackers (more than 50 FPS), including the teacher networks SiamRPN~\cite{li2018high} and SiamFC~\cite{bertinetto2016fully}, Siam-tri \cite{dong2018triplet}, TRACA~\cite{choi2018context}, HP~\cite{dong2018hyperparameter}, Cfnet2~\cite{valmadre2017end}, and fDSST~\cite{danelljan2017discriminative}. The evaluation metrics include both precision and success plots in OPE~\cite{wuobject2015}, where ranks are sorted using precision scores with center error less than 20 pixels and Area-Under-the-Curve (AUC). In Fig.~\ref{fig:res_otb_100}, our DSTrpn outperforms all the other trackers in terms of precision and success plots. As for speed, DSTrpn runs at an extremely high speed of 265 FPS, which is nearly 3$\times$ faster than SiamRPN (90 FPS) and obtains the same (even slightly better) precision and AUC scores. DSTfc runs more than 2$\times$ faster than SiamFC with comparable performance.

		\begin{figure*}
		\centering
		\includegraphics[width = 1\textwidth]{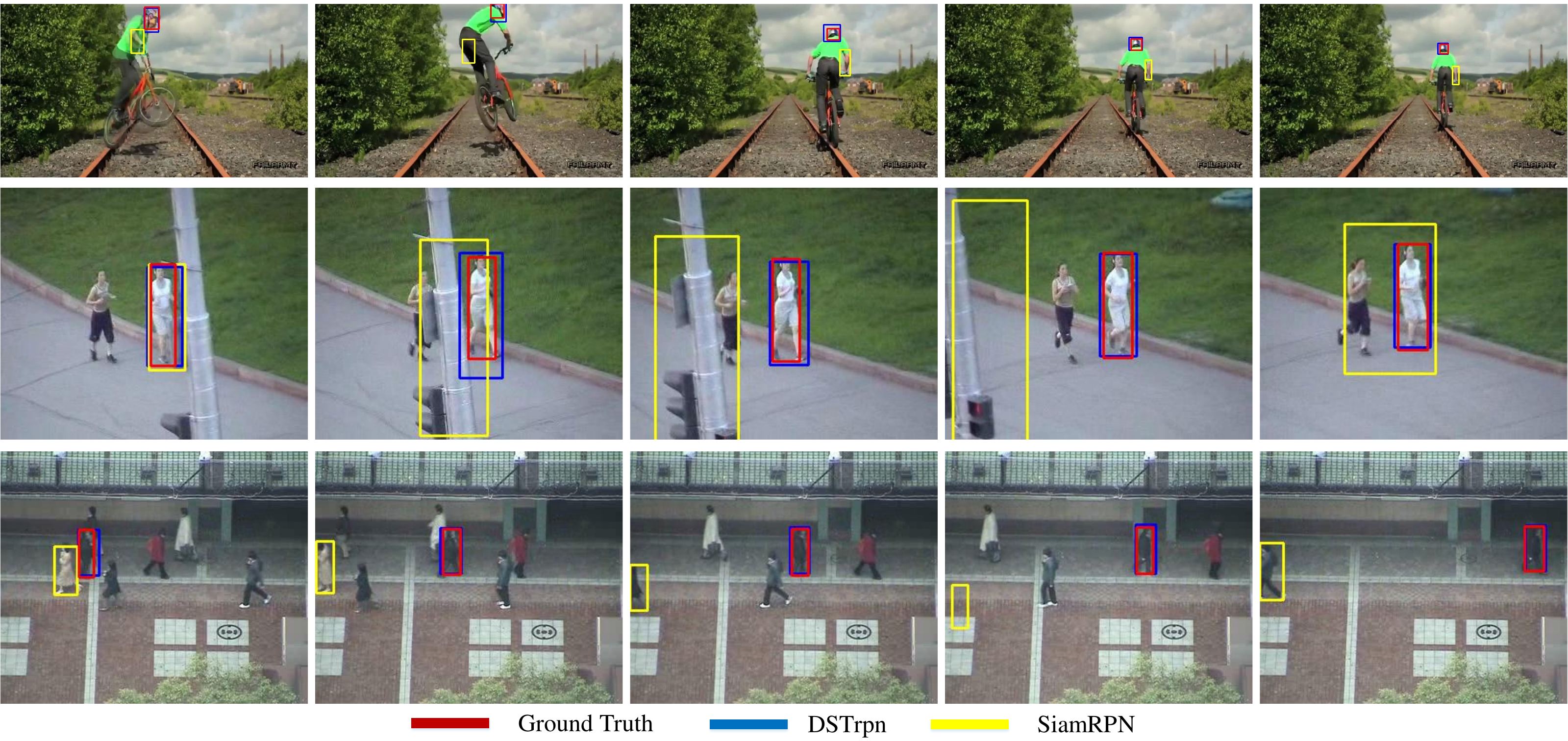}
		\caption{Sample results of SiamRPN and our DSTrpn on OTB-100~\cite{wuobject2015} sequences (Biker, Jogging-2 and Subway).
        On these sequences, our DSTrpn outperforms SiamRPN while running at a much faster speed.}
		\label{fig:quan_res}
	\end{figure*}
	
	\begin{table}
		\small
		\centering
		\resizebox{0.5\textwidth}{!}
		{
			\begin{tabular}{c|ccc|cc|c}
				\hline
				\multirow{2}*{}
				& \multicolumn{3}{c|}{VOT 2019}
				& \multicolumn{2}{c|}{TrackingNet}\\
				\cline{2-6}
				&EAO &A &R &AUC
				&$P$ &FPS\\
				\hline
				ECO~\cite{danelljan2017eco} &/&/&/ &0.554&0.492&8\\
				MDNet~\cite{nam2015learning} &/&/&/ &0.606&0.565&1\\
				DaSiamRPN~\cite{zhu2018distractor} &/&/&/ &0.638&0.591 &160\\
				\hline
				SiamRPN~\cite{li2018high} &0.272&0.582&0.527 &0.675&0.622 &90\\
				SiamFC~\cite{bertinetto2016fully} &0.183&0.511&0.923 &0.573&0.52 &110\\
				\hline
				DSTrpn &0.247 &0.552 &0.637 &0.649&0.589 &265\\
				DSTfc &0.182&0.504&0.923 &0.562&0.512 &230\\
				\hline
			\end{tabular}
		}
		\vspace*{3pt}
		\caption{Results comparison on VOT2019~\cite{Kristan2019a} in terms of EAO, A (Accuracy) and R (Robustness), LaSOT~\cite{fan2019lasot} and TrackingNet~\cite{muller2018trackingnet} in terms of AUC, $P$ (Precision).}
		\label{LaSOT}
	\end{table}

	\begin{figure*}
		\centering
		\includegraphics[width = 1\textwidth]{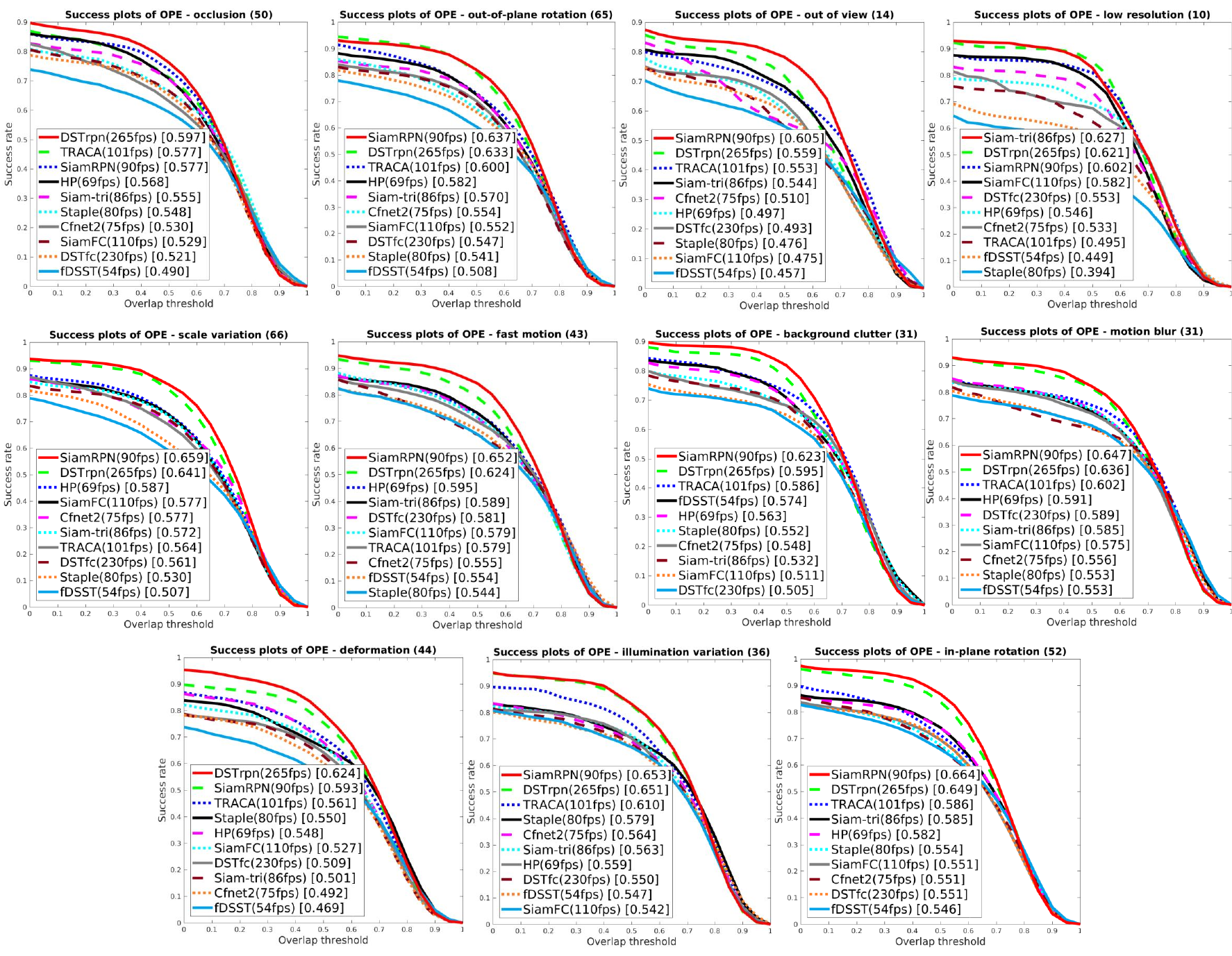}
		\caption{Overlap success plots of OPE with AUC for 11 tracking challenges on OTB-100~\cite{wuobject2015} including: \textit{Illumination Variation (IV), Scale Variation (SV), Occlusion (OCC), Deformation (DEF), Motion Blur (MB), Fast Motion (FM), In-Plane Rotation (IPR), Out-of-Plane Rotation (OPR), Out-of-View (OV), Background Clutter (BC)} and \textit{Low Resolution (LR)}. Our method achieves the best performance.}
		\label{fig:otb_elements}
	\end{figure*}
	
	\noindent\textbf{Results on VOT2019, LaSOT and TrackingNet.} We also conduct extensive experiments on challenging and large-scale datasets, including VOT2019~\cite{Kristan2019a}, LaSOT~\cite{fan2019lasot} and TrackingNet~\cite{muller2018trackingnet}, to evaluate the generalization of our method. On VOT2019, the trackers are ranked by EAO (Expected Average Overlap) while on LaSOT and TrackingNet, OP and DP are used. We compare DaSiamRPN~\cite{zhu2018distractor}, ECO~\cite{danelljan2017eco}, MDNet~\cite{nam2015learning}, and our baselines: SiamRPN~\cite{Kristan2018a} and SiamFC~\cite{bertinetto2016fully}.
	
	As shown in Table~\ref{LaSOT} and Fig.~\ref{fig:res_lasot}, the model size of our DSTrpn (or DSTfc) is further smaller than its teacher model SiamRPN (or SiamFC), while the AUC scores on two large-scale datasets are very close (about 0.02 on DSTrpn) to the teacher. This strongly demonstrates the robustness of the two distilled trackers on long and various videos. {Based on SiamRPN~\cite{li2018high}, DaSiamRPN introduces a distractor-aware model updating strategy, as well as a global detection module.
	Note that, our DSTrpn achieves better performance than DaSiamRPN in both two datasets while maintaining smaller model size and not using any complex strategy or additional module.}
	Via the proposed KD training, our trackers achieve comparable performance with few accuracy losses and much higher speeds on long and challenging videos.

    \noindent\textbf{Results on FaceTracking.}
    {Face tracking is one of the important scenarios for evaluating tracking models. 
    We evaluate our methods on FaceTracking~\cite{qi2020siamese}, the results are shown in Table~\ref{uavdt}. AUC and distance precision are used on both benchmarks like TrackingNet~\cite{muller2018trackingnet}. We compare ECO~\cite{danelljan2017eco}, MDNet~\cite{nam2015learning}, LGT~\cite{qi2020siamese}, and our baselines: SiamRPN~\cite{Kristan2018a} and SiamFC~\cite{bertinetto2016fully}.
    Our distilled trackers obtain comparable or even slightly better performance.
    These results further demonstrate the effectiveness and generalization of the proposed KD method.
    }

	\begin{table}
	\small
	\centering
	{
		\begin{tabular}{c|cc|cc|c}
			\hline
			\multirow{2}*{}
			& \multicolumn{2}{c|}{FaceTracking}\\
			\cline{2-3}
			&AUC &$P$ &FPS\\
			\hline
			ECO~\cite{danelljan2017eco}  &0.538&0.834&8\\
			MDNet~\cite{nam2015learning}  &0.499&0.833&1\\
			LGT~\cite{qi2020siamese}&0.559&0.833&4\\
			\hline
			SiamRPN~\cite{li2018high} &0.453&0.809&90\\
			SiamFC~\cite{bertinetto2016fully}  &0.425&0.694&110\\
			\hline
			DSTrpn  &0.452&0.813&265\\
			DSTfc &0.430&0.698&230\\
			\hline
		\end{tabular}
	}
	\vspace*{3pt}
	\caption{{Comparison on FaceTracking~\cite{qi2020siamese} in terms of AUC and $P$ (Precision).}}
	\label{uavdt}
    \end{table}

	\subsection{Qualitative Evaluation}
	We compare our DSTrpn method with SiamRPN~\cite{li2018high} on several challenging sequences from OTB-100~\cite{wuobject2015} in Fig.~\ref{fig:quan_res}.
{The previous knowledge distillation (KD) work \cite{romero2014fitnets} has revealed that the student model with KD can outperform the teacher in some cases or on some data distributions. Similarly, our student model DSTrpn also achieves better performance on several challenging sequences compared with its teacher SiamRPN. For example, on the Biker sequence (first row in Fig.~\ref{fig:quan_res}),}
	SiamRPN fails to track objects well, whereas our DSTrpn algorithm performs accurately in terms of both precision and overlap. The SiamRPN method gradually loses track of the target due to significant Deformation (DEF) and Fast Motion (FM) in Biker sequence.
On Jogging-2 and Subway, our tracker overcomes Occlution (OCC), Out-of-Plane Rotation (OPR) and Background Clutter (BC), maintaining high tracking accuracy. On all these sequences, our DSTrpn outperforms the teacher while running much more faster. In Fig.~\ref{fig:otb_elements}, the overlap scores of our DSTrpn and other trackers on 11 tracking challenges are shown. Our method achieves the best results on all challenges, while running much faster than other trackers. This reveals the effectiveness of our method for challenging scenes.
    {Further, with the shared knowledge from the other student, our students can even outperform their teachers in certain cases and obtain better overall performance on certain benchmarks, including OTB-100.}
	
	\subsection{Ablation Study}
	\label{sec:siamrpn&siamfc4}
	\begin{table}
		\small
		\centering
		\resizebox{0.49\textwidth}{!}{
			\begin{tabular}{c|c|cccc|c|c}
	 			\hline
				\multicolumn{2}{c|}{}&GT &AH &TS &STR &Precision &AUC\\
				\hline
				\multirow{8}*{SiamRPN}
				&\multirow{7}*{Student1}
				&\checkmark & & & &0.638&0.429\\
				&& & &\checkmark & &0.796&0.586\\
				&&\checkmark & &\checkmark & &0.795&0.579\\
				&& &\checkmark &\checkmark & &0.800&0.591\\
				&& & &\checkmark &\checkmark &0.811&0.608\\
				&&\checkmark & &\checkmark &\checkmark &0.812&0.606\\
				&& &\checkmark &\checkmark &\checkmark &0.825&0.624\\
				\cline{2-8}
				&Teacher &/&/&/&/&0.853&0.643\\
				\hline
				\multirow{6}*{SiamFC}
				&\multirow{5}*{Student1}
				&\checkmark & & & &0.707&0.523\\
				&& & &\checkmark & &0.711&0.535\\
				&&\checkmark & &\checkmark & &0.710&0.531\\
				&& & &\checkmark &\checkmark &0.742&0.548\\
				&&\checkmark & &\checkmark &\checkmark &0.741&0.557\\
				\cline{2-8}
				&Teacher &/&/&/&/&0.772&0.581\\
	 			\hline
			\end{tabular}
		}
		\vspace*{3pt}
		\caption{Results for different combinations of GT, TS, AH and STR in terms of precision and AUC on OTB-100~\cite{wuobject2015}.}
		\label{ablation1}
	\end{table}
	
	\noindent\textbf{Knowledge Transfer Components.} The teacher-student knowledge transfer consists of three components: (i) AH loss, (ii) TS loss, and (iii) STR loss. We conduct an extensive ablation study by implementing a number of variants using different combinations, including (1) GT: simply using hard labels, (2) TS, (3) GT+TS, (4) AH+TS, (5) TS+STR, (6) GT+TS+STR, and (7) AH+TS+STR (the full knowledge transfer method).
	Table~\ref{ablation1} shows our results on SiamFC and SiamRPN. For SiamRPN, we can see that the GT without any proposed loss degrades dramatically compared with the teacher, due to the lack of a pre-trained backbone. When using the TS loss to train student, we observe a significant improvement in terms of precision ($15.8\%$) and AUC ($15.7\%$). However, directly combining GT and TS (GT+TS) could be suboptimal due to over-fitting. By replacing GT with AH, AH+TS further boosts the performance for both metrics. Finally, by adding the STR loss, the model (AH+TS+STR) is able to close the gap between the teacher and student, outperforming other variants.
	In addition to the results shown in Table~\ref{ablation1}, we do two more studies for the final model (AH+TS+STR) to further demonstrate the impact of the newly proposed weighting strategy in STR and distillation loss (AH+TS). In the first study, we remove the weighting strategy and achieved the worse Precision (0.814) and lower AUC (0.609). In the second one, we replace our distillation loss with the bounded loss~\cite{chen2017learning} and got a reduced performance of 0.817 and 0.612 in terms of the Precision and AUC, respectively.
	SiamFC only employs the classification loss, so GT is equal to AH and we use GT here. Results show that the gaps are narrower than SiamRPN but improvements are still obvious. These results clearly demonstrate the effectiveness of each component.
		
		\begin{table}
		\small
		\centering
		\resizebox{0.48\textwidth}{!}{
			\begin{tabular}{c|c|c|c|c|c|c}
				\hline
				\multicolumn{2}{c|}{}  &{NOKD} &TSKD &TSsKD &Size &FPS\\
				\hline
				\multirow{3}*{SiamRPN}  &Student1 &0.429 &0.624 &0.646 &19.7M &265\\
				\cline{2-7}
				&Student2 &0.630 &0.641 &0.644 &90.6M &160\\
				\cline{2-7}
				&Teacher &0.642 &/ &/ &361.8M &90\\
				\hline
				\multirow{3}*{SiamFC}  &Student1 &0.523 &0.557 &0.573 &0.7M &230\\
				\cline{2-7}
				&Student2 &0.566 &0.576 &0.579 &2.4M &165\\
				\cline{2-7}
				&Teacher &0.581 &/ &/ &9.4M &110\\
				\hline
			\end{tabular}
		}
		\vspace*{3pt}
		\caption{Ablation experiments of different learning mechanisms (NOKD, KD, TSsKD) in terms of AUC on OTB-100~\cite{wuobject2015}.}
		\label{ablation2}
	\end{table}

\begin{figure}[t]
	\centering
	\includegraphics[width = 0.49 \textwidth]{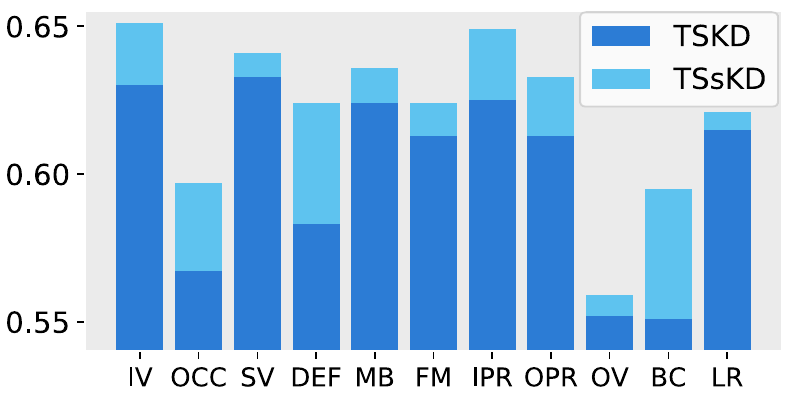}
	\caption{The AUC scores of different learning mechanisms on 11 challenging attributes of OTB-100~\cite{wuobject2015}.}
	\label{fig:attributes}
\end{figure}

	\noindent\textbf{Different Learning Mechanisms.} To evaluate our TSsKD model, we also conduct an ablation study on different learning mechanisms 
	: (i) NOKD: trained with hard labels, (ii) TSKD: our tracking-specific teacher-student knowledge distillation (transfer) and (iii) TSsKD. ``Student1'' and ``Student2'' represent the ``dim'' and ``intelligent'' student, respectively.
	Students are trained following different paradigms and results can be seen in Table~\ref{ablation2}. With KD, all students are improved. Moreover, with the knowledge sharing in our TSsKD, the ``dim'' SiamRPN student gets a performance improvement of $2.2\%$ in terms of AUC. The ``dim'' SiamFC student gets a $1.6\%$ improvement.
	{More specifically, the shared knowledge from the ``intelligent'' student is effective and provides promising improvement on all challenging tracking cases. As shown in Fig.~\ref{fig:attributes}, TSsKD outperforms TSKD on all 11 challenging attributes of OTB-100~\cite{wuobject2015}. 
	In terms of DEF (\textit{Deformation}), and BC (\textit{Background Clutter}), knowledge sharing obtains a significant improvement of more than $4\%$.
	The TSsKD model obviously enhances the robustness of ``dim'' student.}
	Besides, the ``intelligent'' SiamRPN and SiamFC students get slight improvements ($0.3\%$) as well. Fusing the knowledge from the teacher, ground-truth and ``intelligent'' student, the ``dim'' SiamRPN student obtains the best performance.

		\begin{figure}
		\centering
		\subfigure[]{
			\label{fig:student_number-siamrpn}
			\includegraphics[width = .224 \textwidth]{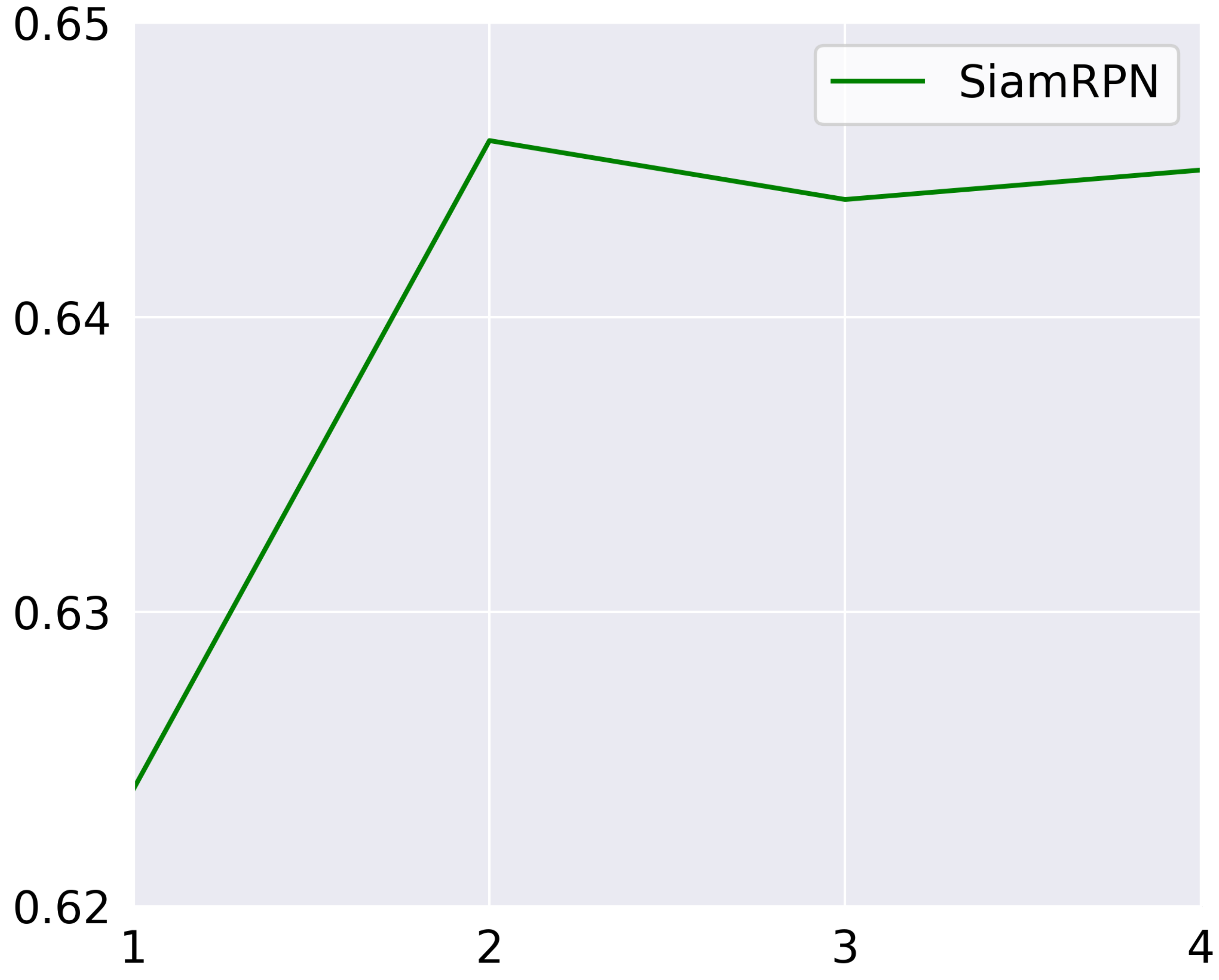}
		}
		\subfigure[]{
			\label{fig:student_number-siamfc}
			\includegraphics[width = .224 \textwidth]{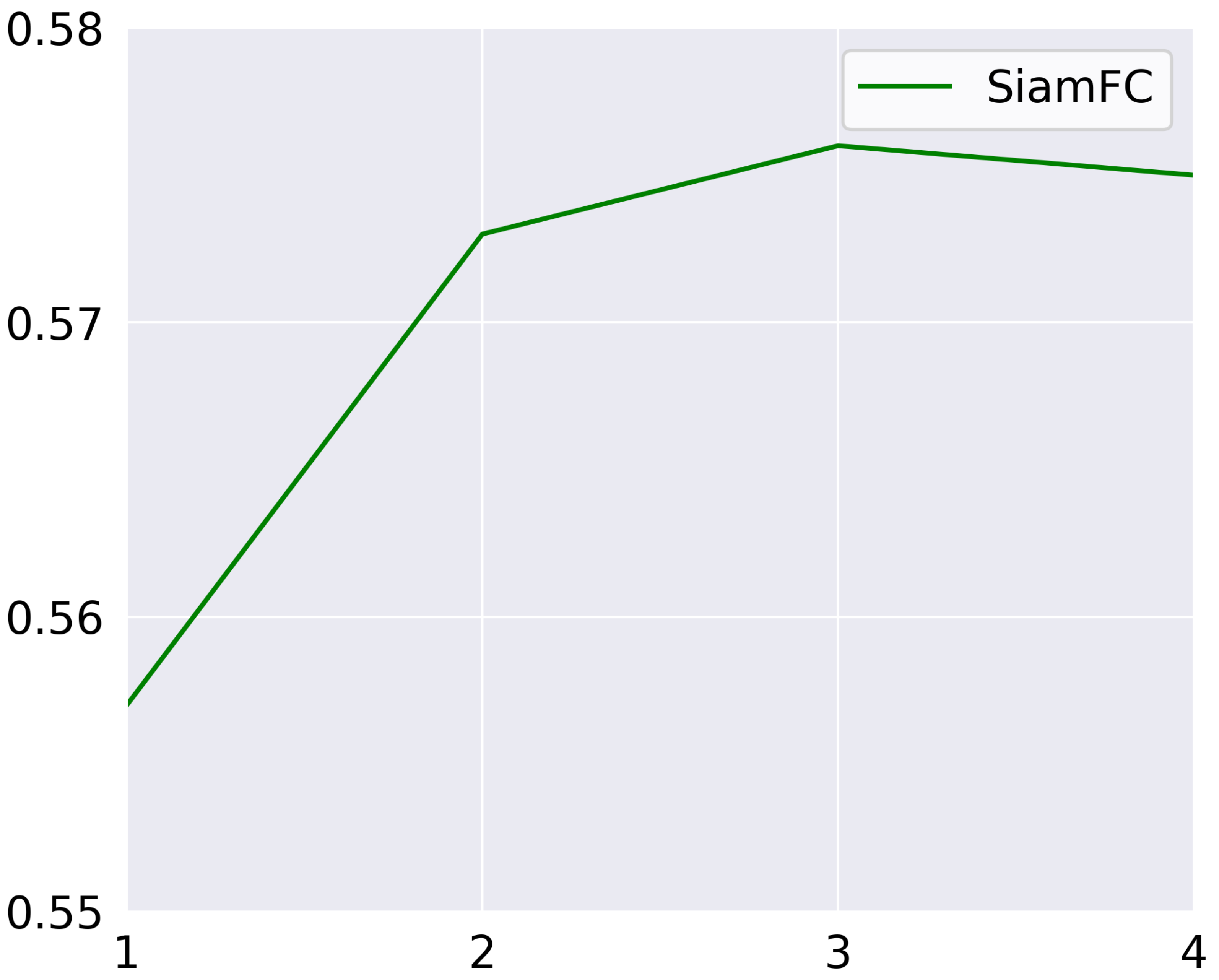}
		}
		\caption{Performance of (a) DSTrpn and (b) DSTfc on OTB-100~\cite{wuobject2015} with different student numbers in terms of AUC.}
		\label{fig:student_number}
	\end{figure}
	
    \noindent\textbf{Experiments on More Students.}
	Our TSsKD model can be extended to more students. Given $n$ students s1, s2, ..., sn, the objective function for si is as follows:
	\begin{equation}
	\label{eq:KS-loss1}
	L^{\text{KD}}_{\text{si}} =  L^{\text{KT}}_{\text{si}} + \frac{1}{n}  \sum\nolimits_{j=1}^n \beta_{ij} \sigma (\text{s1}) L^{\text{KS}}(\text{si}||\text{sj}).
	\end{equation}
	Here $\beta_{ij}$ is the discount factor between si and sj considering their different reliabilities. For example, in our case of two students, $\beta_{12}$ = 1 and $\beta_{21}$ = 0.5. We conduct an experiment using different numbers of student and report the results in Fig.~\ref{fig:student_number}. Students are generated by reducing the number of convolutional channels to a scale (0.4, 0.45, 0.5, 0.55). In our case, since our ``dim'' students already achieve a performance similar to the teacher's with only one ``intelligent'' student, more students do not bring significant improvements.
	
	\noindent\textbf{Sensibility Evaluation.}
	To further evaluate the robustness of the proposed KD methods, we select several models generated during the DRL process to evaluate our method's sensibility to model selection. In this experiment, we select several models and Table~\ref{sensibility} reports their tracking performance on OTB~\cite{wuobject2015}. We can see that their performances are very comparable to the teachers', verifying our method's
	robustness to the architectures of the searched networks.
	
	\noindent\textbf{Grayscale.}
	{To evaluate the impact of grayscales on the training of student networks, we conduct an experiment to train them with different scales of gray images and test them on OTB dataset~\cite{wuobject2015}. As shown in Table~\ref{grayscale}, 15\% to 25\% is a suitable setting for performance, and too many or too few gray images in the training will cause a performance drop.}

	\noindent\textbf{CPU Speed.}
	To provide a more practical speed comparison of different networks, we also test them on an Intel(R) Xeon(R) 2.20GHz CPU. In this computation-constrained running environment, our DSTrpn and DSTfc can run at a speed of 20 and 30 FPS, respectively, which is close to real-time speed. SiamRPN and SiamFC only achieve 8 and 12 FPS. These results clearly demonstrate our merits.

	\begin{table}
	\centering
	{
		\begin{tabular}{c|c|c|c}
			\hline
			DRL Iteration times &{35} &{40} &{45} \\
			\hline
			SiamRPN  &0.645  &0.646 (\textbf{DSTrpn})  &0.642\\
			\hline
			SiamFC &0.573 (\textbf{DSTfc}) &0.580  &0.576\\
			\hline
		\end{tabular}
	}
	\vspace*{3pt}
	\caption{AUC score of searched students in different DRL iterations, tested on the OTB~\cite{wuobject2015}.}
	\label{sensibility}
	\end{table}
	
		\begin{table}
	\centering
	{
		\begin{tabular}{c|c|c|c|c|c}
			\hline
			Grayscale &{0\%} &{5\%} &{15\%} &{25\%} &{35\%}\\
			\hline
			DSTrpn  &0.621  &0.631  &0.643 &0.646 &0.640\\
			\hline
			DSTfc &0.565 &0.570  &0.574 &0.573 &0.569\\
			\hline
		\end{tabular}
	}

	\caption{{AUC score of students trained with different scales of gray images, tested on the OTB~\cite{wuobject2015}.}}
	\label{grayscale}
	\end{table}
	
	\begin{table}[tp]
		\centering
		\resizebox{0.5\textwidth}{!}
		{
			\begin{tabular}{c|ccc|cc|cc|c}
				\hline
				\multirow{2}*{}
				& \multicolumn{3}{c|}{VOT 2019}
				& \multicolumn{2}{c|}{LaSOT}
				& \multicolumn{2}{c|}{TrackingNet}\\
				\cline{2-8}
				&EAO &A &R &AUC &$P_{norm}$ &AUC
				&$P$ &FPS\\
				\hline
				SiamRPN++ &0.287&0.596&0.472 &0.496&0.568 &0.733&0.694 &35\\
				SiamRPN++r34 (w/o) &0.270&0.585&0.515 &0.464&0.548 &0.690&0.651 &50\\
				SiamRPN++r18 (w/o)&0.255&0.586&0.552 &0.443&0.520 &0.672&0.617 &75\\
				\hline
				SiamRPN++r34 (w)&0.288&0.604&0.484 &0.472&0.562 &0.699&0.657 &50\\
				SiamRPN++r18 (w)&0.271&0.588&0.517 &0.465&0.544 &0.676&0.623 &75\\
				\hline
			\end{tabular}
		}
		\vspace*{3pt}
		\caption{Results of different trackers trained with(w)/without(w/o) TSsKD. ``r34'' and ``r18'' denote trackers using ResNet34 and ResNet18 as their backbone, respectively.}
		\label{rpn++}
	\end{table}
	
	\subsection{Extended Experiments on SiamRPN++}
	\label{sec:siamrpn&siamfc5}
	The backbone network of SiamRPN++ is pre-trained on ImageNet for image labeling. Two sibling convolutional layers are attached to the stride-reduced ResNet-50 to perform proposal classification and bounding
	box regression with five anchors. Three randomly initialized 1x1 convolutional layers are attached to conv3, conv4, conv5 for reducing the feature dimension to 256. The whole network is trained with stochastic gradient descent (SGD) on the four datasets as with SiamRPN. A warmup learning rate of 0.001 is used for the first five epochs to train the RPN branches. For the last 15 epochs, the whole network is end-to-end trained with the learning rate exponentially decayed	from 0.005 to 0.0005. A weight decay of 0.0005 and momentum
	of 0.9 are used. The training loss is the sum of the classification
	loss and the standard smooth L1 loss for regression.
	
	{The backbone of SiamRPN++ is deep and it is difficult to obtain a reliable performance without pre-training it on ImageNet, so it is not suitable to use the same reinforcement learning method as SiamRPN. On the other side, there are many smaller classic networks with guaranteed performance (such as ResNet18, ResNet34~\cite{he2016deep}), which can be selected as our student models.}
	In this part, we first train the original SiamRPN++ tracker (with ResNet50 as the backbone).
	Then, two SiamRPN++ models with a pre-trained ResNet34 and ResNet18 backbone are trained simultaneously as the students in our TSsKD.
{As shown in Table~\ref{rpn++}, our TSsKD can further improve the SOTA SiamRPN++ trackers with small backbones (ResNet34 or ResNet18).}
All training settings are kept the same as in \cite{li2019siamrpn++}. We can see that, with the proposed knowledge distillation method, both trackers are improved significantly on all benchmarks.	
	
	\section{Conclusion}
	This paper proposed a new Distilled Siamese Tracker (DST) framework to learn small, fast and accurate trackers from larger Siamese trackers. This framework is built upon a teacher-students knowledge distillation model that includes two types of knowledge transfer: 1) knowledge transfer from teacher to students by a tracking-specific distillation strategy; 2) mutual learning between students in a knowledge sharing manner. The theoretical analysis and extensive empirical evaluations on two Siamese trackers have clearly demonstrated the generality and effectiveness of the proposed DST. Specifically, for the SOTA SiamRPN, the distilled tracker achieved a high compression rate, ran at an extremely high speed, and obtained a similar performance as the teacher. Thus, we believe such a distillation method could be used for improving many SOTA deep trackers for practical tracking tasks.

\section{Acknowledgements}
\small{This work was supported in part by the FDCT grant SKL-IOTSC(UM)-2021-2023,
and Start-up Research Grant (SRG) of University of Macau (SRG2022-00023-IOTSC).
The first three authors contributed equally to this paper.}

	

\begin{thebibliography}{10}
		
		\bibitem{ashok2017n2n}
		A.~Ashok, N.~Rhinehart, F.~Beainy, and K.~M. Kitani.
		\newblock N2n learning: network to network compression via policy gradient reinforcement learning.
		\newblock In {\em the International Conference on Learning Representations (ICLR)}, 2018.
		
		\bibitem{ba2014deep}
		J.~Ba and R.~Caruana.
		\newblock Do deep nets really need to be deep?
		\newblock In {\em Conference on Neural Information Processing Systems (NeurIPS)}, pp. 2654-2662, 2014.
		
		\bibitem{bertinetto2016fully}
		L.~Bertinetto, J.~Valmadre, J.~F. Henriques, A.~Vedaldi, and P.~H. Torr.
		\newblock Fully-convolutional siamese networks for object tracking.
		\newblock In {\em European Conference on Computer Vision (ECCV) Workshop}, 2016.
		
		\bibitem{bertinetto2016learning}
		L.~Bertinetto, J.~F. Henriques, J.~Valmadre, P.~H. Torr and A.~Vedaldi.
		\newblock Learning feed-forward one-shot learners.
		\newblock In {\em Conference on Neural Information Processing Systems (NeurIPS)}, pp. 523-531, 2016.
		
		\bibitem{bolme2010visual}
		Bolme, David S and Beveridge, J Ross and Draper, Bruce A and Lui, Yui Man.
		\newblock Visual object tracking using adaptive correlation filters.
		\newblock In {\em IEEE Conference on Computer Vision and Pattern Recognition (CVPR)}, pp. 2544-2550, 2010.
		
		\bibitem{bucilu2006model}
		C.~Bucilu, R.~Caruana, and A.~Niculescu-Mizil.
		\newblock Model compression.
		\newblock In {\em ACM SIGKDD conference}, pp. 535¨C541, 2006.
		
		\bibitem{chen2017learning}
		G.~Chen, W.~Choi, X.~Yu, T.~Han, and M.~Chandraker.
		\newblock Learning efficient object detection models with knowledge distillation.
		\newblock In {\em Conference on Neural Information Processing Systems (NeurIPS)}, pp. 742-751, 2017.
		
		\bibitem{choi2018context}
		J.~Choi, H.~J. Chang, T.~Fischer, S.~Yun, K.~Lee, J.~Jeong, Y.~Demiris, and J.~Y. Choi.
		\newblock Context-aware deep feature compression for high-speed visual tracking.
		\newblock In {\em IEEE Conference on Computer Vision and Pattern Recognition (CVPR)}, pp. 479-488, 2018.
		
		\bibitem{choi2019deep}
        J.~Choi, J.~Kwon and K.~Lee.
        \newblock Deep meta learning for real-time target-aware visual tracking.
        \newblock In {\em IEEE Conference on International Conference on Computer Vision (ICCV)}, pp. 911-920, 2019.
		
		\bibitem{courbariaux2015binaryconnect}
		M.~Courbariaux, Y.~Bengio and J.~P.~David.
		\newblock Binaryconnect: Training deep neural networks with binary weights during propagations.
		\newblock In {\em Conference on Neural Information Processing Systems (NeurIPS)}, pp. 3123-3131, 2015.
		
		\bibitem{courbariaux2016binarized}
		M.~Courbariaux, I.~Hubara, D.~Soudry, R.~El-Yaniv and Y.~Bengio.
		\newblock Binarized neural networks: Training deep neural networks with weights and activations constrained to+ 1 or-1.
		\newblock In {\em arXiv preprint arXiv:1602.02830}, 2016.
		
		\bibitem{czarnecki2017sobolev}
		W.~M. Czarnecki, S.~Osindero, M.~Jaderberg, G.~Swirszcz, and R.~Pascanu.
		\newblock Sobolev training for neural networks.
		\newblock In {\em Conference on Neural Information Processing Systems (NeurIPS)}, pp. 4278-4287, 2017.
		
		\bibitem{dai2019visual}
		Dai, Kenan and Wang, Dong and Lu, Huchuan and Sun, Chong and Li, Jianhua.
		\newblock Visual Tracking via Adaptive Spatially-Regularized Correlation Filters.
		\newblock In {\em IEEE Conference on International Conference on Computer Vision (ICCV)}, pp. 4670-4679, 2019.
		
		\bibitem{danelljan2016beyond}
		M.~Danelljan, A.~Robinson, G.~Bhat, F.~S. Khan, M.~Felsberg, et~al.
		\newblock Beyond correlation filters: Learning continuous convolution operators for visual tracking.
		\newblock In {\em European Conference on Computer Vision (ECCV)}, pp. 472-488, 2016.
		
		\bibitem{danelljan2017eco}
		M.~Danelljan, G.~Bhat, F.~S. Khan, M.~Felsberg, et~al.
		\newblock Eco: Efficient convolution operators for tracking.
		\newblock In {\em IEEE Conference on Computer Vision and Pattern Recognition (CVPR)}, pp. 6931-6939, 2017.
		
		\bibitem{danelljan2014accurate}
		M.~Danelljan, G.~H{\"a}ger, F.~Khan, and M.~Felsberg.
		\newblock Accurate scale estimation for robust visual tracking.
		\newblock In {\em the British Machine Vision Conference (BMVC)}, 2014.
		
		\bibitem{danelljan2014adaptive}
		M.~Danelljan, G.~Bhat, F.~S. Khan, M.~Felsberg, et~al.
		\newblock Adaptive color attributes for real-time visual tracking.
		\newblock In {\em IEEE Conference on Computer Vision and Pattern Recognition (CVPR)}, pp. 1090-1097, 2014.
		
		\bibitem{danelljan2017discriminative}
		M.~Danelljan, G.~H{\"a}ger, F.~S. Khan, and M.~Felsberg.
		\newblock Discriminative scale space tracking.
		\newblock {\em IEEE Transactions on Pattern Analysis and Machine Intelligence}, 39(8):1561--1575, 2017.
		
		\bibitem{danelljan2015learning}
		M.~Danelljan, G.~Hager, F.~Shahbaz~Khan, and M.~Felsberg.
		\newblock Learning spatially regularized correlation filters for visual tracking.
		\newblock In {\em IEEE Conference on International Conference on Computer Vision (ICCV)}, pp. 4310-4318, 2015.
		
		\bibitem{dong2018triplet}
		X.~Dong and J.~Shen.
		\newblock Triplet loss in siamese network for object tracking.
		\newblock In {\em European Conference on Computer Vision (ECCV)}, pp. 472-488, 2018.
		
		\bibitem{fan2019lasot}
		H.~Fan, L.~Lin, F.~Yang, P.~Chu, G.~Deng, S.~Yu, H.~Bai, Y.~Xu, C.~Liao, and H.~Ling.
		\newblock Lasot: A high-quality benchmark for large-scale single object tracking.
		\newblock In {\em IEEE Conference on Computer Vision and Pattern Recognition (CVPR)}, pp. 5374-5383, 2019.
		
		\bibitem{fan2019siamese}
		H.~Fan, H.~Ling.
		\newblock Siamese cascaded region proposal networks for real-time visual tracking.
		\newblock In {\em IEEE Conference on Computer Vision and Pattern Recognition (CVPR)}, pp. 5374-5383, 2019

		\bibitem{dong2018hyperparameter}
		X.~Dong, J.~Shen, W.~Wang, Y.~Liu, L.~Shao, and F.~Porikli.
		\newblock Hyperparameter optimization for tracking with continuous deep Q-learning.
		\newblock In {\em IEEE Conference on Computer Vision and Pattern Recognition (CVPR)}, pp. 518-527, 2018.
		
		\bibitem{furlanello2018born}
		T.~Furlanello, Z.~C. Lipton, M.~Tschannen, L.~Itti, and A.~Anandkumar.
		\newblock Born again neural networks.
		\newblock In {\em the International Conference on Machine Learning (ICML)}, 2018.
		
		\bibitem{gao2019graph}
        J.~Gao, T.~Zhang and C.~Xu.
        \newblock Graph convolutional tracking.
        \newblock In {\em IEEE Conference on Computer Vision and Pattern Recognition (CVPR)}, pp. 4649-4659, 2019.
		
		\bibitem{gupta2015deep}
		S.~Gupta, A.~Agrawal, K.~Gopalakrishnan and P.~Narayanan.
		\newblock Deep learning with limited numerical precision.
		\newblock In {\em the International Conference on Machine Learning (ICML)}, pp. 1737-1746, 2015.
		
		\bibitem{han2015deep}
		Deep compression: Compressing deep neural networks with pruning, trained quantization and huffman coding.
		\newblock S.~Han, H.~Mao and D.~William J.
		\newblock In {\em arXiv preprint arXiv:1510.00149}, 2015.

        \bibitem{shen2020hierarchical}
        J. Shen, X. Tang, X. Dong, and L. Shao,
        \newblock Visual object tracking by hierarchical attention Siamese network,
        \newblock {\em IEEE Trans. on Cybernetics}, vol. 50, no. 7, pp. 3068-3080, 2020.
		
		\bibitem{he2016deep}
		K.~He, X.~Zhang, S.~Ren, and J.~Sun.
		\newblock Deep residual learning for image recognition.
		\newblock In {\em IEEE Conference on Computer Vision and Pattern Recognition (CVPR)}, pages 770--778, 2016.
		
		\bibitem{henriques2012exploiting}
		J.~F.~Henriques, C.~Rui, M.~Pedro and B.~Jorge.
		\newblock Exploiting the circulant structure of tracking-by-detection with kernels.
		\newblock In {\em European Conference on Computer Vision (ECCV)}, pp. 702-715, 2012.
		
		\bibitem{henriques2014high}
		J.~F.~Henriques, C.~Rui, M.~Pedro and B.~Jorge.
		\newblock High-speed tracking with kernelized correlation filters.
		\newblock In {\em IEEE Transactions on Pattern Analysis and Machine Intelligence}, vol. 37, no. 3, pp. 583-596, 2015.
		
		\bibitem{hinton2015distilling}
		G.~Hinton, O.~Vinyals, and J.~Dean.
		\newblock Distilling the knowledge in a neural network.
		\newblock In {\em Conference on Neural Information Processing Systems (NeurIPS) Workshop}, 2014.
		
		\bibitem{huang2019bridging}
        L.~Huang, X.~Zhao and K.~Huang.
        \newblock Bridging the gap between detection and tracking: A unified approach.
        \newblock In {\em IEEE Conference on International Conference on Computer Vision (ICCV)}, pp. 3998-4008, 2019.
		
		\bibitem{kart2019object}
		K.~Ugur, L.~Alan, K.~Matej, K.~Joni-Kristian and M.~Jiri.
		\newblock Object Tracking by Reconstruction with View-Specific Discriminative Correlation Filters.
		\newblock In {\em IEEE Conference on Computer Vision and Pattern Recognition (CVPR)}, pp. 1339-1348, 2019.
		
		\bibitem{kiani2013multi}
		K.~Galoogahi, Hamed, S.~Terence and L.~Simon.
		\newblock Multi-channel correlation filters.
		\newblock In {\em IEEE Conference on International Conference on Computer Vision (ICCV)}, 2013.
		
		\bibitem{kiani2017learning}
		K.~Galoogahi, Hamed, F.~Ashton and L.~Simon.
		\newblock Learning background-aware correlation filters for visual tracking.
		\newblock In {\em IEEE Conference on International Conference on Computer Vision (ICCV)}, pp. 1144-1152, 2017.
		
		\bibitem{kim2018paraphrasing}
		J.~Kim, S.~Park and N.~Kwak
		\newblock Paraphrasing complex network: Network compression via factor transfer.
		\newblock In {\em Conference on Neural Information Processing Systems (NeurIPS)}, pp. 2765-2774, 2018.
		
		\bibitem{Kristan2018a}
		M.~Kristan, A.~Leonardis, J.~Matas, M.~Felsberg, R.~Pfugfelder, L.~C. Zajc,
		T.~Vojir, G.~Bhat, A.~Lukezic, A.~Eldesokey, G.~Fernandez, and et~al.
		\newblock The sixth visual object tracking vot2018 challenge results.
		\newblock In {\em European Conference on Computer Vision (ECCV) Workshop}, 2018.
		
		\bibitem{Kristan2019a}
		M.~Kristan, A.~Leonardis, J.~Matas, M.~Felsberg, R.~Pfugfelder, L.~C. Zajc,
		T.~Vojir, G.~Bhat, A.~Lukezic, A.~Eldesokey, G.~Fernandez, and et~al.
		\newblock The seventh visual object tracking vot2019 challenge results.
		\newblock In {\em IEEE Conference on International Conference on Computer Vision (ICCV) workshop}, 2019.
		
		\bibitem{krizhevsky2012imagenet}
		A.~Krizhevsky, I.~Sutskever, and G.~E. Hinton.
		\newblock Imagenet classification with deep convolutional neural networks.
		\newblock In {\em Conference on Neural Information Processing Systems (NeurIPS)}, pp. 1106-1114, 2012.
		
        \bibitem{han2021fuse}
        W. Han, X. Dong, F. S. Khan, L. Shao, J. Shen,
        \newblock Learning To Fuse Asymmetric Feature Maps in Siamese Trackers,
        \newblock In {\em IEEE Conference on Computer Vision and Pattern Recognition (CVPR)}, pp. 16570-16580, 2021.

		\bibitem{li2014scale}
		Y.~Li and J.~Zhu.
		\newblock A scale adaptive kernel correlation filter tracker with feature integration.
		\newblock In {\em European Conference on Computer Vision (ECCV) Workshop}, 2014.
		
		\bibitem{li2018high}
		B.~Li, J.~Yan, W.~Wu, Z.~Zhu, and X.~Hu.
		\newblock High performance visual tracking with siamese region proposal network.
		\newblock In {\em IEEE Conference on Computer Vision and Pattern Recognition (CVPR)}, pp. 8971-8980, 2018.
		
         \bibitem{Dong2021hyper}
         X. Dong, J. Shen, W. Wang, L. Shao, H. Ling, and F. Porikli,
         \newblock Dynamical Hyperparameter Optimization via Deep Reinforcement Learning in Tracking,
         \newblock IEEE Trans. on Pattern Analysis and Machine Intelligence, vol. 43, no. 5, pp. 1515-1529, 2021.
		
		\bibitem{li2019siamrpn++}
		B.~Li, W.~Wu, Q.~Wang, F.~Zhang, J.~Xing, and J.~Yan.
		\newblock Siamrpn++: Evolution of siamese visual tracking with very deep networks.
		\newblock In {\em IEEE Conference on Computer Vision and Pattern Recognition (CVPR)}, pp. 4282-4291, 2019.
		
		\bibitem{li2019target}
        X.~Li, C.~Ma, B.~Wu, Z.~He and M.~Yang.
        \newblock Target-aware deep tracking.
        \newblock In {\em IEEE Conference on Computer Vision and Pattern Recognition (CVPR)}, pp. 1369-1378, 2019.

		\bibitem{lin2014microsoft}
		T.-Y. Lin, M.~Maire, S.~Belongie, J.~Hays, P.~Perona, D.~Ramanan,
		P.~Doll{\'a}r, and C.~L. Zitnick.
		\newblock Microsoft coco: Common objects in context.
		\newblock In {\em European Conference on Computer Vision (ECCV)}, pp. 740-755, 2014.
		
		\bibitem{liu2019structured}
		Y.~Liu, K.~Chen, C.~Liu, Z.~Qin, Z.~Luo and J.~Wang
		\newblock Structured Knowledge Distillation for Semantic Segmentation.
		\newblock In {\em IEEE Conference on Computer Vision and Pattern Recognition (CVPR)}, pp. 2604-2613, 2019.
		
		\bibitem{lopez2015unifying}
		D.~Lopez-Paz, L.~Bottou, B.~Sch{\"o}lkopf, and V.~Vapnik.
		\newblock Unifying distillation and privileged information.
		\newblock In {\em International Conference on Learning Representations (ICLR)}, 2016.
		
		\bibitem{ma2015long}
		C.~Ma, X.~Yang, C.~Zhang, M.~Yang.
		\newblock Long-term correlation tracking.
		\newblock In {\em IEEE Conference on Computer Vision and Pattern Recognition (CVPR)}, pp. 5388-5396, 2015.
		
		\bibitem{muller2018trackingnet}
		M.~Muller, A.~Bibi, S.~Giancola, S.~Alsubaihi, and B.~Ghanem.
		\newblock Trackingnet: A large-scale dataset and benchmark for object tracking in the wild.
		\newblock In {\em European Conference on Computer Vision (ECCV)}, pp. 310-327, 2018.
		
		\bibitem{naresh2013correlation}
		N.~Boddeti, V., Kanade, T., and Vijaya Kumar, B. V. K.
		\newblock Correlation filters for object alignment.
		\newblock In {\em IEEE Conference on Computer Vision and Pattern Recognition (CVPR)}, 2013.
		
		\bibitem{nam2015learning}
		H.~Nam and B.~Han.
		\newblock Learning multi-domain convolutional neural networks for visual tracking.
		\newblock In {\em IEEE Conference on Computer Vision and Pattern Recognition (CVPR)}, pp. 4293-4302, 2016.
		
		\bibitem{possegger2015defense}
		H.~Possegger, T.~Mauthner, and H.~Bischof.
		\newblock In defense of color-based model-free tracking.
		\newblock In {\em IEEE Conference on Computer Vision and Pattern Recognition (CVPR)}, pp. 2113-2120, 2015.
		
		\bibitem{qi2018hedging}
		Y.~Qi, S.~Zhang, L.~Qin, Q.~Huang, H.~Yao, J.~Lim and M.~Yang.
		\newblock Hedging deep features for visual tracking.
		\newblock {\em IEEE Trans. on Pattern Analysis and Machine Intelligence}, vol. 41, no. 5, pp. 1116-1130, 2019.
		
		\bibitem{qi2020siamese}
		Y.~Qi, S.~Zhang, J.~Feng, H.~Zhou, D.~Tao and X.~Li.
		\newblock Siamese local and global networks for robust face tracking.
		\newblock In {\em IEEE Transactions on Image Processing}, vol. 29, pp. 9152-9164, 2020.
		
		\bibitem{rastegari2016xnor}
		M.~Rastegari, V.~Ordonez, J.~Redmon and A.~Farhadi.
		\newblock Xnor-net: Imagenet classification using binary convolutional neural networks.
		\newblock In {\em European Conference on Computer Vision (ECCV)}, pp. 525-542, 2016.
		
		\bibitem{real2017youtube}
		E.~Real, J.~Shlens, S.~Mazzocchi, X.~Pan, and V.~Vanhoucke.
		\newblock Youtube-boundingboxes: A large high-precision human-annotated data set for object detection in video.
		\newblock In {\em IEEE Conference on Computer Vision and Pattern Recognition (CVPR)}, pp. 7464-7473, 2017.
		
		\bibitem{ren2015faster}
		S.~Ren, K.~He, R.~Girshick, and J.~Sun.
		\newblock Faster r-cnn: Towards real-time object detection with region proposal networks.
		\newblock In {\em Conference on Neural Information Processing Systems (NeurIPS)}, pp. 91-99, 2015.
		
		\bibitem{romero2014fitnets}
		A.~Romero, N.~Ballas, S.~E. Kahou, A.~Chassang, C.~Gatta, and Y.~Bengio.
		\newblock Fitnets: Hints for thin deep nets.
		\newblock In {\em International Conference on Learning Representations (ICLR)}, 2015.
		
		\bibitem{ILSVRC15}
		O.~Russakovsky, J.~Deng, H.~Su, J.~Krause, S.~Satheesh, S.~Ma, Z.~Huang,
		A.~Karpathy, A.~Khosla, M.~Bernstein, A.~C. Berg, and L.~Fei-Fei.
		\newblock {ImageNet Large Scale Visual Recognition Challenge}.
		\newblock {\em International Journal of Computer Vision}, vol. 115, no. 3, pp. 211-252, 2015.
		
		\bibitem{sadowski2015deep}
		P.~Sadowski, J.~Collado, D.~Whiteson, and P.~Baldi.
		\newblock Deep learning, dark knowledge, and dark matter.
		\newblock In {\em Conference on Neural Information Processing Systems (NeurIPS) Workshop}, 2015.
		
		\bibitem{simonyan2014very}
		S.~Karen and Z.~Andrew.
		\newblock Very deep convolutional networks for large-scale image recognition.
		\newblock In {\em arXiv preprint arXiv:1409.1556}, 2014.
		
		\bibitem{srinivas2015data}
		S.~Srinivas and R V.~Babu.
		\newblock Data-free parameter pruning for deep neural networks.
		\newblock In {\em arXiv preprint arXiv:1507.06149}, 2015.
		
		\bibitem{sun2019roi}
		Y.~Sun, C.~Sun, D.~Wang, Y.~He and H.~Lu.
		\newblock ROI Pooled Correlation Filters for Visual Tracking.
		\newblock In {\em IEEE Conference on Computer Vision and Pattern Recognition (CVPR)}, pp. 5783-5791, 2019.
		
        \bibitem{Dong2019Quadruplet}
        \newblock X. Dong, J. Shen, D. Wu, K. Guo, X. Jin, F. Porikli,
        \newblock Quadruplet Network With One-Shot Learning for Fast Visual Object Tracking,
        \newblock {\em IEEE Trans. on Image Processing}, vol. 28, no. 7, pp. 3516-3527, 2019.
		
		\bibitem{tao2016siamese}
		R.~Tao, E.~Gavves, and A.~W. Smeulders.
		\newblock Siamese instance search for tracking.
		\newblock In {\em IEEE Conference on Computer Vision and Pattern Recognition (CVPR)}, pp. 1420-1429, 2016.
		
		\bibitem{urban2016deep}
		G.~Urban, K.~J. Geras, S.~E. Kahou, O.~Aslan, S.~Wang, R.~Caruana, A.~Mohamed,
		M.~Philipose, and M.~Richardson.
		\newblock Do deep convolutional nets really need to be deep and convolutional?
		\newblock In {\em International Conference on Learning Representations (ICLR)}, 2017.
		
		\bibitem{valmadre2017end}
		J.~Valmadre, L.~Bertinetto, J.~F. Henriques, A.~Vedaldi, and P.~H. Torr.
		\newblock End-to-end representation learning for correlation filter based tracking.
		\newblock In {\em IEEE Conference on Computer Vision and Pattern Recognition (CVPR)}, pp.5000-5008, 2017.
		
		\bibitem{vapnik1998statistical}
		V.~Vapnik.
		\newblock Statistical learning theory, 1998.
		
		\bibitem{wang2013learning}
		N.~Wang and D.~Yeung.
		\newblock Learning a deep compact image representation for visual tracking.
		\newblock In {\em Conference on Neural Information Processing Systems (NeurIPS)}, pp. 809-817, 2013.
		
		\bibitem{wang2015transferring}
		N.~Wang, S.~Li, A.~Gupta, and D.-Y. Yeung.
		\newblock Transferring rich feature hierarchies for robust visual tracking.
		\newblock {\em arXiv preprint arXiv:1501.04587}, 2015.
		
		\bibitem{wang2015understanding}
		N.~Wang, J.~Shi, D.-Y. Yeung, and J.~Jia.
		\newblock Understanding and diagnosing visual tracking systems.
		\newblock In {\em IEEE Conference on International Conference on Computer Vision (ICCV)}, pp. 3101-3109, 2015.

        \bibitem{Liang2020local}
        Z. Liang, and J. Shen,
        \newblock Local Semantic Siamese Networks for Fast Tracking,
        \newblock {\em IEEE Trans. on Image Processing}, vol. 29, no. 3351-3364, 2020
		
		\bibitem{wang2020real}
        N.~Wang, W.~Zhou, Y.~Song, C.~Ma and H.~Li.
        \newblock Real-Time Correlation Tracking Via Joint Model Compression and Transfer.
        \newblock {\em IEEE Transactions on Image Processing }, vol. 29, pp. 6123--6135, 2020.
		
		\bibitem{williams1992simple}
		R.~J. Williams.
		\newblock Simple statistical gradient-following algorithms for connectionist reinforcement learning.
		\newblock {\em Machine Learning}, 8(3-4):229--256, 1992.
		
		\bibitem{wuobject2015}
		W.~Yi, L.~Jongwoo, and M.-H. Yang.
		\newblock Object tracking benchmark.
		\newblock {\em IEEE Transactions on Pattern Analysis and Machine Intelligence}, vol. 37, no. 9, pp. 1834--1848, 2015.

        \bibitem{Lu2020Shrinkage}
        X. Lu, C. Ma, J. Shen, X. Yang, I. Reid, and M.-H. Yang,
        \newblock Deep Object Tracking with Shrinkage Loss,
        \newblock {\em IEEE Trans. on Pattern Analysis and Machine Intelligence}, doi://10.1109/TPAMI.2020.3041332, 2020

        \bibitem{yang2020release}
		Y.~Yang, G.~Li, Y.~Qi and Q.~Huang.
		\newblock Release the Power of Online-Training for Robust Visual Tracking.
		\newblock In {\em AAAI Conference on Artificial Intelligence}, pp. 12645-12652, 2020.

		\bibitem{yang2020roam}
		T.~Yang, P.~Xu, R.~Chai and A.~B. Chan.
		\newblock ROAM: Recurrently Optimizing Tracking Model.
		\newblock In {\em IEEE Conference on Computer Vision and Pattern Recognition (CVPR)}, pp. 6717-6726, 2020.
		
		\bibitem{zagoruyko2016paying}
		S.~Zagoruyko and N.~Komodakis.
		\newblock Paying more attention to attention: Improving the performance of convolutional neural networks via attention transfer.
		\newblock In {\em International Conference on Learning Representations (ICLR)}, 2016.
		
		\bibitem{Zhang2014MEEM}
		J.~Zhang, S.~Ma, and S.~Sclaroff.
		\newblock Meem: Robust tracking via multiple experts using entropy minimization.
		\newblock In {\em European Conference on Computer Vision (ECCV)}, pp. 188-203, 2014.
		
		\bibitem{zhang2018deep}
		Y.~Zhang, T.~Xiang, T.~M. Hospedales, and H.~Lu.
		\newblock Deep mutual learning.
		\newblock In {\em IEEE Conference on Computer Vision and Pattern Recognition (CVPR)}, pp. 4320-4328, 2018.
		
		\bibitem{zhang2019deeper}
		Z.~Zhang and H.~Peng.
		\newblock Deeper and wider siamese networks for real-time visual tracking.
		\newblock In {\em IEEE Conference on Computer Vision and Pattern Recognition (CVPR)}, pp. 4591-4600, 2019.
		
		\bibitem{zhang2020ocean}
		Z.~Zhang and H.~Peng.
		\newblock Ocean: Object-aware anchor-free tracking.
		\newblock In {\em European Conference on Computer Vision (ECCV)}, pp. 771-787, 2020.
		
		\bibitem{zhu2018distractor}
		Z.~Zhu, Q.~Wang, B.~Li, W.~Wu, J.~Yan, and W.~Hu.
		\newblock Distractor-aware siamese networks for visual object tracking.
		\newblock In {\em European Conference on Computer Vision (ECCV)}, pp. 103-119, 2018.

	\end{thebibliography}
	%
	{\small
		\bibliographystyle{IEEEtran}
	
}	
	\vfill
	%
\end{document}